\documentclass[11pt]{article}

\usepackage[preprint]{acl}

\usepackage{times}
\usepackage{latexsym}

\usepackage[T1]{fontenc}

\usepackage[utf8]{inputenc}

\usepackage{microtype}

\usepackage{inconsolata}

\usepackage{graphicx}
\usepackage{booktabs}
\usepackage{multirow} 
\usepackage{float}
\usepackage{amsmath}
\usepackage{enumitem}
\usepackage{xcolor}
\usepackage{dblfloatfix}
\usepackage{subcaption}
\usepackage{textcomp}
\usepackage{xspace}
\usepackage{makecell}

\usepackage{todonotes}

\definecolor{myblue}{HTML}{1155cc}
\definecolor{myred}{HTML}{990000}
\definecolor{mycolor}{HTML}{afdcff}

\newcommand{\wbox}[1]{\fcolorbox{white}{white}{#1}}
\newcommand{\bbox}[1]{\fcolorbox{mycolor}{mycolor}{#1}}

\newcommand{\hippo}{\textsc{HIPPO}\xspace}

\newcommand{\twitter}{\includegraphics[height=1.8ex]{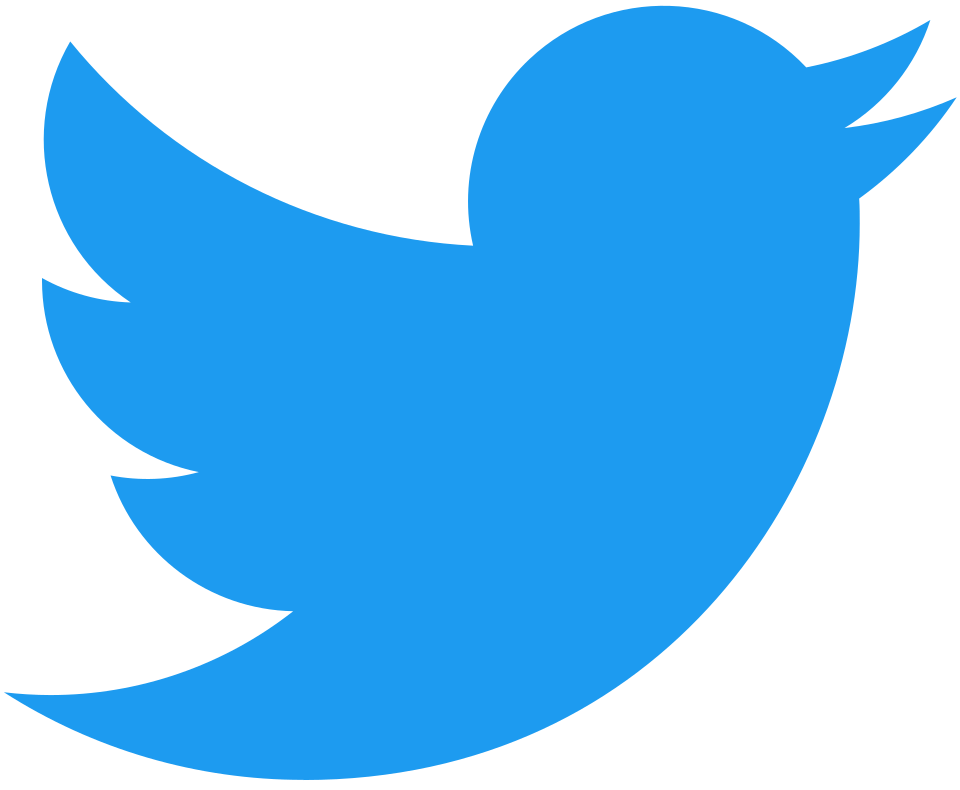}}
\newcommand{\reddit}{\includegraphics[height=1.8ex]{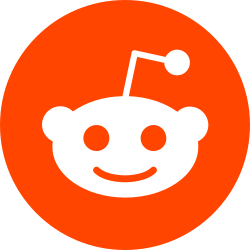}}
\newcommand{\youtube}{\includegraphics[height=1.8ex]{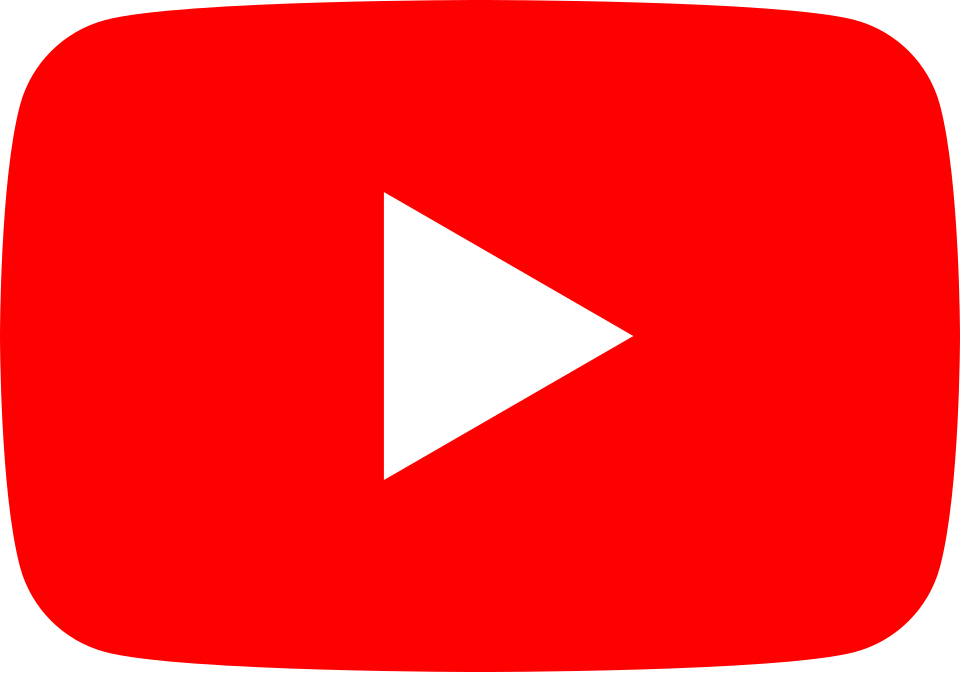}}
\newcommand{\facebook}{\includegraphics[height=1.8ex]{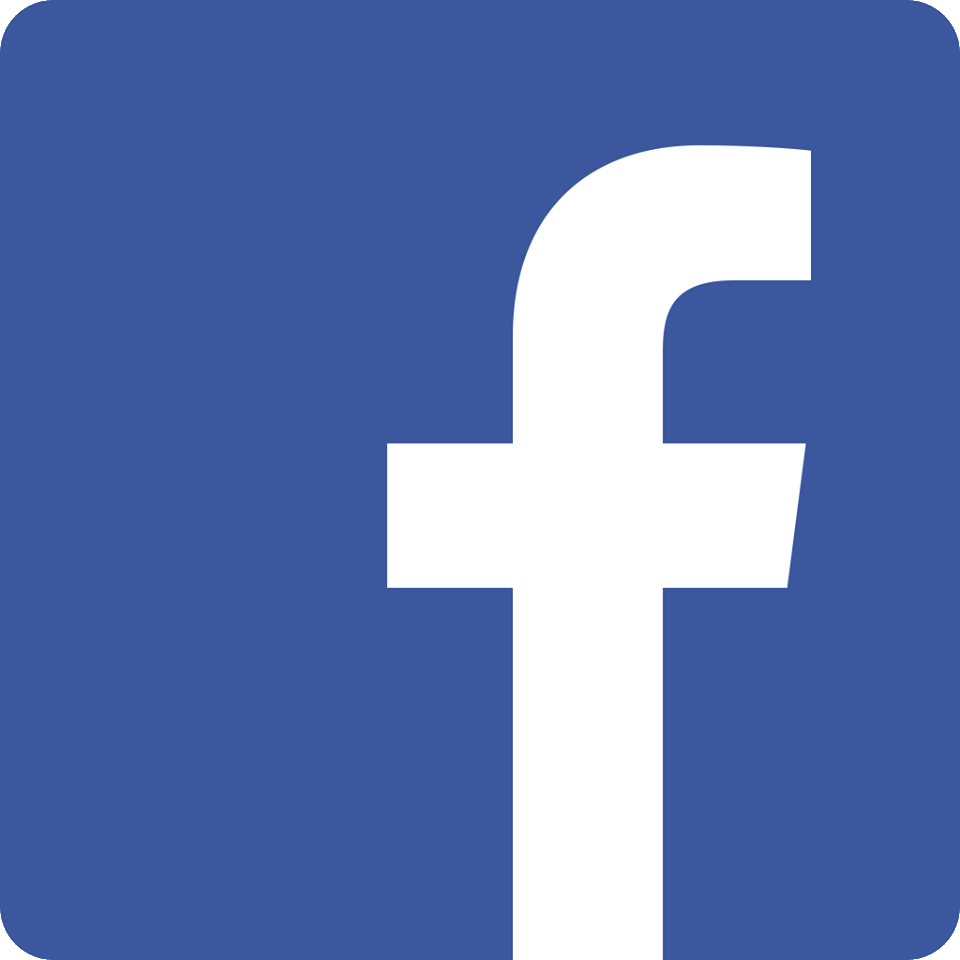}}
\newcommand{\gab}{\includegraphics[height=1.8ex]{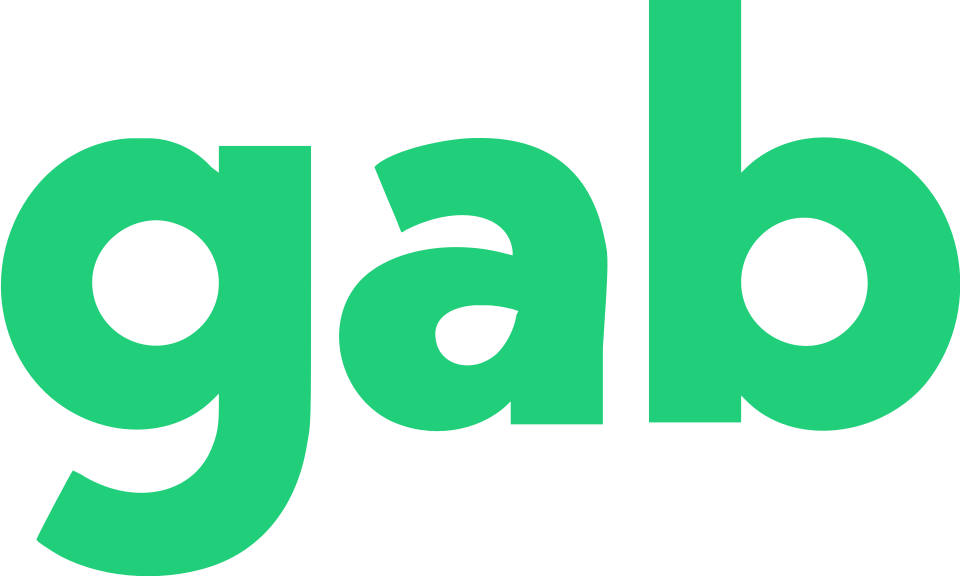}}
\newcommand{\wiki}{\includegraphics[height=1.8ex]{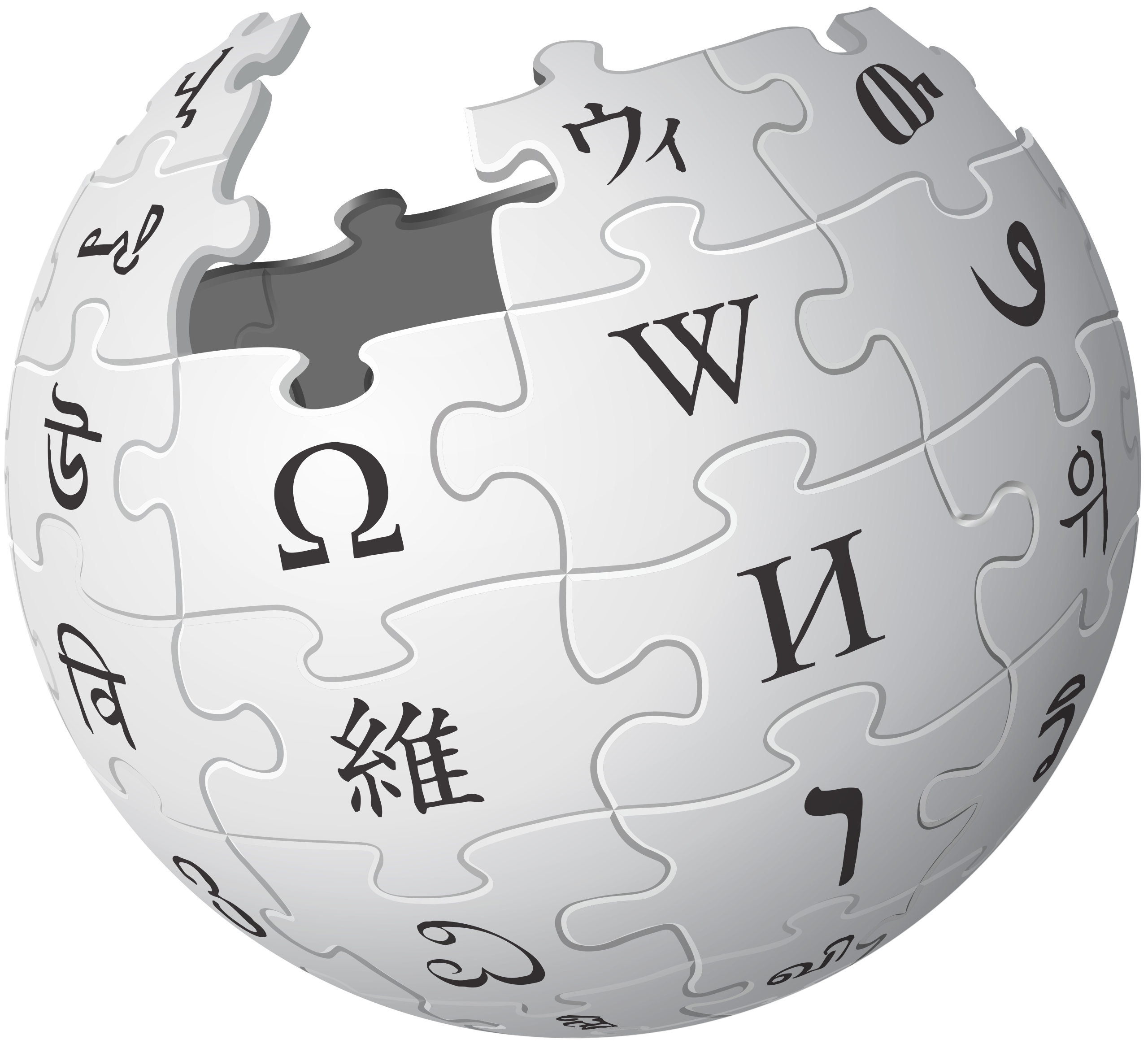}}
\title{From Specialization to Generalization: \\ Instruction-tuned LLMs for Robust Harmful Content Mitigation}

\author{Lukas Edman$^{1,2}$ \qquad Daryna Dementieva$^{1,2}$ \qquad Alexander Fraser$^{1,2,3}$ \vspace{.2cm}\\ 
$^{1}$School of Computation, Information and Technology, TU Munich \\
$^{2}$Munich Center for Machine Learning \\ 
$^{3}$Munich Data Science Institute \\ 
\vspace{.1cm} {\tt \small lukas.edman@tum.de, daryna.dementieva@tum.de}
}

\begin{document}
\maketitle
\begin{abstract}

Large language models (LLMs) demonstrate impressive performance across a wide range of general NLP tasks; however, their effectiveness in sensitive domains, such as hate speech detection, remains less clear. Prior studies comparing prompted LLMs with state-of-the-art encoder-based models (e.g., BERT variants~\cite{roy2023probing,donmez-etal-2024-please}) have shown only marginal gains, suggesting that LLMs may not excel in hate speech detection or mitigation. In this work, we revisit this question through the lens of instruction tuning. By thoroughly unifying 36 English hate speech datasets spanning multiple labeling schemes, we fine-tune a generalist LLM, based on Qwen3~\cite{qwen3technicalreport}, specifically for hate speech mitigation. Our results demonstrate not only state-of-the-art performance on in-domain benchmarks but also substantial improvements in cross-domain and cross-lingual generalization---areas where encoder-based specialist classifiers often struggle. 

\end{abstract}

\section{Introduction}


Hate speech detection has been heavily studied, leading to the release of numerous datasets covering a wide spectrum of tasks, domains, languages, and annotation schemes~\cite{mathew2021hatexplain,mandl2019overview,jigsaw-toxic-comment-classification-challenge} (see Table~\ref{tab:datasets}). Despite this diversity, progress in building \textit{generalizable} hate speech models, which can cover a large variety of tasks related to hate speech as well as perform well on unseen domains, has been limited. Existing research largely treats each dataset or subtask in isolation, and little effort has been made to unify many available heterogeneous datasets to develop task-agnostic models with more robust performance across varied hate speech scenarios.

Consequently, task-specific encoder-based models (e.g., BERT variants) remain the dominant approach~\cite{wang-etal-2020-galileo,DBLP:conf/fire/SaiS20}, as prompting large language models (LLMs) has not consistently yielded significantly superior results~\cite{roy2023probing,donmez-etal-2024-please}. At the same time, instruction tuning, one of the most promising methods for aligning LLMs to specialized domains, has not yet been systematically explored for hate speech mitigation.

\begin{figure}[t!]
    \centering
    \includegraphics[width=0.5\textwidth]{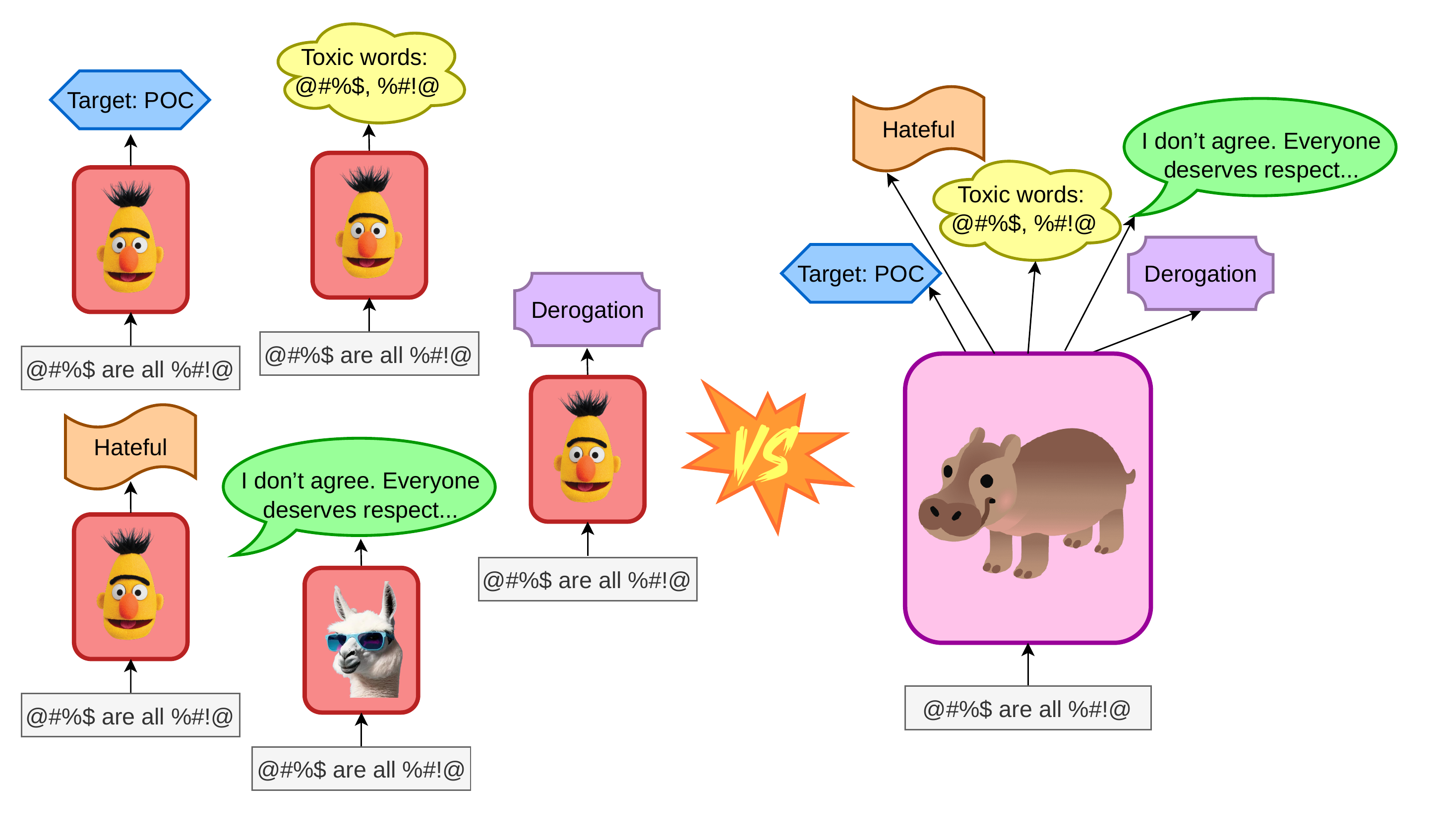}
    \caption{In this work, we would like to close the gap in understanding if instruction-tuned general-purpose LLMs (i.e., HIPPO, right) are more robust and precise detectors of hate speech than specially-tuned LMs (i.e., BERT and Llama, left).}
    \label{fig:intro}
\end{figure}

To close this gap, firstly, we consolidate \textbf{36 heterogeneous datasets} in English into a single, instruction-formatted corpus spanning over 600k examples and 61 subtasks. Using this unified training dataset, we instruction-tune an LLM and name the resulting model \textbf{HIPPO} (\textbf{H}armful \textbf{I}nput, \textbf{P}ositive and \textbf{P}roductive \textbf{O}utput, see Figure~\ref{fig:intro}). We evaluate HIPPO across in-domain, out-of-domain, and cross-lingual settings. Our study aims to answer the following research questions:

\newcommand{\rqone}{How does our instruction-tuned LLM perform compared to strong task-specific baselines for hate speech tasks in English?}
\newcommand{\rqtwo}{Does our instruction-tuned LLM outperform zero- or few-shot LLMs prompting for hate speech tasks?}
\newcommand{\rqthree}{Can our instruction-tuned LLM generalize to previously unseen hate speech domains and labels?}
\newcommand{\rqfour}{To what extent can our English-only instruction-tuning dataset support cross-lingual generalization in LLMs?}

\begin{description}
    \itemsep=-3pt
    \item[\textbf{RQ1}:] \rqone
    \item[\textbf{RQ2}:] \rqtwo
    \item[\textbf{RQ3}:] \rqthree
    \item[\textbf{RQ4}:] \rqfour
\end{description}

Our results show that the instruction-tuned model achieves competitive, and often better, performance compared to best-performing encoder-based systems, while also exhibiting cross-task generalization. With a smaller version of the dataset to prime the model, we also see potentially higher cross-lingual performance, especially for generative tasks.

We release our code used for fine-tuning and the best-performing models online for public use.\footnote{\url{https://github.com/TUM-NLP/HIPPO}}



\begin{table*}[!htp]\centering
\scriptsize
\begin{tabular}{lrrrrrrr}\toprule
Dataset &Attribution &Train &Test &Domain &Label Type &(Sub)Tasks \\\midrule
CAD &\citet{vidgen2021introducing} &13423 & &\reddit &Multilabel &1 \\
CAT-LARGE &\citet{pavlopoulos2020toxicity} &19870 & &\wiki &Binary &1 \\
CONAN &\citet{chung-etal-2019-conan} &408 & &Synthetic &Generation &1 \\
ConvAbuse &\citet{cercas-curry-etal-2021-convabuse} &4013 & &Chatbot &Hierarchical &4 \\
DialoCONAN &\citet{bonaldi-etal-2022-human} &1299 & &Synthetic &Generation &1 \\
EAP &\citet{vidgen2020detecting} &19989 & &\twitter &Multiclass &1 \\
ETHOS &\citet{mollas2022ethos} &998 & &\reddit~\youtube &Hierarchical &4 \\
FoxCom &\citet{gao2017detecting} &1525 & &Fox News &Binary &1 \\
GHC &\citet{kennedy2022introducing} &20610 & &\gab &Hierarchical &3 \\
HSOL &\citet{davidson2017automated} &24771 & &\twitter &Multiclass &1 \\
ImplicitHate &\citet{elsherief-etal-2021-latent} &6358 & &\twitter &Multiclass &1 \\
Intervene &\citet{qian2019benchmark} &16628 & &\reddit~\gab &Generation &1 \\
LargeScaleAbuse &\citet{founta2018large} &44768 & &\twitter &Multiclass &1 \\
LargeScaleXDomain &\citet{toraman-etal-2022-large} &68188 & &\twitter &Multiclass &1 \\
MeasuringHate &\citet{sachdeva-etal-2022-measuring,kennedy2020constructing} &39511 & &\twitter~\reddit~\youtube &Multiclass &1 \\
Multitarget-CONAN &\citet{fanton-2021-human} &4055 & &Synthetic &Generation &2 \\
NewsHate &\citet{salminen2018anatomy} &3215 & &\facebook~\youtube &Hierarchical &2 \\
ReligiousHate &\citet{ramponi2022addressing} &5746 & &\twitter &Hierarchical &5 \\
SlurCorpus &\citet{kurrek-etal-2020-towards} &39944 & &\reddit &Multiclass &1 \\
Stormfront &\citet{de2018hate} &10521 & &Stormfront &Binary &1 \\
SWAD &\citet{pamungkas2020you} &2567 & &\twitter &Binary &1 \\
ToxiCR &\citet{sarker2023automated} &12767 & &Code review &Binary &1 \\
TwitterCRT &\citet{waseem-hovy:2016:N16-2} &12727 & &\twitter &Binary &1 \\
TwitterExpert &\citet{waseem:2016:NLPandCSS} &6908 & &\twitter &Multiclass &1 \\
TwitterSA &\citet{toosi2019twitterhate} &25825 & &\twitter &Binary &1 \\
USElect &\citet{grimminger-klinger-2021-hate} &2400 & &\twitter &Binary &1 \\ 
AMI18 &\citet{fersini2018overview} &3998 &1000 &\twitter &Hierarchical &3 \\ 
EDOS &\citet{kirkSemEval2023} &13998 &2000 &\reddit~\gab &Hierarchical &3 \\
HASOC19 &\citet{mandl2019overview} &5817 &1153 &\twitter~\facebook &Hierarchical &3 \\
HatEval19 &\citet{basile2019semeval} &8992 &2971 &\twitter &Hierarchical &3 \\
HateXplain &\citet{mathew2021hatexplain} &15348 &1924 &\twitter~\gab &Hierarchical &3 \\
Jigsaw &\citet{jigsaw-toxic-comment-classification-challenge} &159529 &63978 &CC &Multilabel &1 \\
OffensEval20 &\citet{zampieri2019predicting} &13187 &5309 &\twitter &Hierarchical &3 \\
ParaDetox &\citet{logacheva-etal-2022-paradetox} &11927 &600 &\twitter~\reddit~CC &Generation &1 \\
ToxicSpans &\citet{pavlopoulos-etal-2022-acl,pavlopoulos-etal-2021-semeval} &7931 &2000 &CC &Generation &1 \\ \midrule
Total & &649761 &80935 & & &61 \\ 

\bottomrule
\end{tabular}
\caption{List of datasets used and the number of examples, domain, and type of labels. Domains include Twitter (\twitter), Reddit (\reddit), Gab (\gab), Facebook (\facebook), Wikipedia (\wiki), YouTube (\youtube), and Civil Comments (CC), among others.}
\label{tab:datasets}
\end{table*}

\section{Background}

\paragraph{Hate Speech Definitions}
While a lot of work has been published in the domain of hate speech detection, very few computer science works mention explicitly what definition of hate or abusive speech they adopt in their studies for annotation or automatic processing~\cite{DBLP:journals/csur/AroraNHSNDZDBBA24,rizwan-etal-2025-hateprism}. Thus, firstly, we specify which definitions we use in this study.

Hate speech mitigation lies under the term \textit{digital violence}---an umbrella term that refers to words or actions that cause harm to an individual or a community.
Our study focuses on digital violence, more specifically, expressed in a textual form. The study by \citet{lewandowska2023annotation} categorizes harmful content as \textit{offensive speeches}, including 17 sub-categories like \textit{taboo}, \textit{insulting}, \textit{hate speech}, \textit{harassment}, and \textit{toxic}.

We adopt the definition of \textit{hate speech} as abusive language targeting specific groups \citep{rottger-etal-2021-hatecheck}, and \textit{toxic speech} as texts that contain vulgar or profane language~\cite{costa2022no,logacheva-etal-2022-paradetox}, but are not necessarily abusive or hateful. For many other labels that were then used in the instruction prompts, we refer to the original definitions of the datasets.

\paragraph{Hate Speech Mitigation Datasets}
Over the years, dozens of hate speech detection datasets have been introduced, reflecting a wide range of annotation strategies, domains, and languages~\cite{DBLP:journals/lre/PolettoBSBP21}. The substantial efforts of the NLP community in this area are documented in the Hate Speech Data Catalog, which systematically tracks the development of hate speech resources \cite{vidgen2020directions}.\footnote{\href{[https://hatespeechdata.com}{https://hatespeechdata.com}}

Significant effort has been devoted to developing diverse hate speech detection and mitigation datasets, particularly for English. Common labeling schemes range from \textit{binary} classification (e.g., hate vs. non-hate, toxic vs. non-toxic) to \textit{multiclass} labels (e.g., hate speech vs. offensive language vs. neutral) and even \textit{hierarchical} frameworks that further specify the subtype of hate once identified. One of the earliest curated resources, AMI18~\cite{fersini2018overview}, was derived from Twitter and focused specifically on misogynistic content with a hierarchical labeling strategy. Then, many more datasets were gathered from various social networks like Twitter~\cite{zampieri2019predicting,sachdeva-etal-2022-measuring}, Gab~\cite{mathew2021hatexplain,kirkSemEval2023}, or Reddit~\cite{vidgen2021introducing,qian2019benchmark}. One of the most massive toxic speech classification datasets, Jigsaw~\cite{jigsaw-toxic-comment-classification-challenge}, is based on Civil Comments and includes over 150k labeled samples.

Another notable dataset, HateCheck \cite{rottger-etal-2021-hatecheck}, uses functionality tests to isolate specific cases in which hate speech manifests, allowing researchers to diagnose systematic model failures that may be obscured by aggregate benchmark metrics.

Beyond hate content \textit{classification} datasets, several research directions have emerged that focus on more proactive hate speech mitigation through text \textit{generation}. In particular, to reduce the publication of toxic or profane language---for example, in efforts to promote safer online environments for children---\textit{text detoxification}, which aims to transform toxic text into neutral, has been proposed~\cite{atwell-etal-2022-appdia,dementieva-etal-2025-multilingual}. For more severe cases of hate speech, approaches based on \textit{counter-speech}, incorporating arguments and proactive dialogue, have also been explored~\cite{yu-etal-2022-hate,bonaldi-etal-2022-human}.

Together with English, many hate speech detection resources were created for other languages, for example, for Arabic and French~\citep{ousidhoum2019multilingual}, Spanish~\citep{basile2019semeval}, Italian~\citep{DBLP:conf/clic-it/CorazzaMCTV19}, Portuguese~\citep{fortuna-etal-2019-hierarchically}, German~\citep{DBLP:journals/corr/abs-1910-07518}, among others. 

\paragraph{Hate Speech Supersets}
With many datasets already created for English and other languages, several works have attempted to unify them into a superset for ease of use.
\citet{antypas-camacho-collados-2023-robust} made such an attempt with only \textit{binary} labels 13 English hate speech detection datasets. Then, they performed a cross-dataset performance study utilizing mostly BERT-like or smaller LMs. 
Finally, MetaHate \cite{piot2024metahate} aggregates 36 English datasets into one superset. These datasets either had binarized labels or multiclass labels that were subsequently binarized. This contrasts from our approach, where we do not substantially change the content of any labels, and additionally include generative tasks which are not trivial or possible to binarize.

\paragraph{Hate Speech Detection with Transformer-based Encoders}

Across numerous hate speech detection benchmarks, both in English and in other languages, state-of-the-art performance is predominantly driven by Transformer-based encoder models such as BERT~\cite{devlin-etal-2019-bert} and RoBERTa~\cite{DBLP:journals/corr/abs-1907-11692}, including their multilingual variants. To further enhance domain-specific performance, HateBERT~\cite{caselli-etal-2021-hatebert} was introduced, extending BERT through additional Masked Language Modeling on one million posts from banned Reddit communities.

Correspondingly, for datasets such as AMI18 and HASOC19, the strongest results were achieved using BERT and mBERT-based systems~\cite{muti-barron-cedeno-2022-checkpoint, mishra20193idiots}. In the OffensEval20 shared task, the winning solution employed a RoBERTa-based architecture~\cite{wiedemann-etal-2020-uhh}. Likewise, for toxicity detection in the Jigsaw dataset, one of the top-performing and most widely adopted models~\cite{logacheva-etal-2022-paradetox} is a fine-tuned RoBERTa. Overall, BERT-style encoders continue to dominate performance in the hate speech detection domain.

\paragraph{Hate Speech Detection with LLMs}
With LLMs' appearance, especially with additional human and safety alignment~\cite{DBLP:journals/tist/NaveedKQSAUABM25}, many experiments have been done to exploit such models for hate speech detection and automatic moderation. 

\citet{zhu2023can} reports low agreement between LLM outputs and human labels, while \citet{li2024hot} finds that LLMs are more reliable at identifying non-hateful content. \citet{huang2023chatgpt} studies LLM-generated explanations for implicit hate, and \citet{roy2023probing} shows that adding target-specific context improves prompting performance. \citet{donmez-etal-2024-please} evaluate a range of open- and closed-source LLMs prompting for hate speech and microaggression detection, showing that some models perform competitively. More advanced prompting techniques, like prompting with hate speech definitions or chain-of-thought prompting, showed promising results for real-world hate speech moderation~\cite{guo2023investigation}. At the same time, prompting LLMs can lead to various undesirable and unstable outputs like hallucinations or refusal to process an input sample~\cite{DBLP:journals/air/Huang25}.

Several attempts have been made to fine-tune LLMs for hate speech detection tasks. \citet{DBLP:journals/corr/abs-2405-01577} tuned several 1B models with LoRA on two datasets. A comparison between prompting and fine-tuning of LLMs for sexism detection was performed in \cite{pan2024comparing}. Finally, \citet{DBLP:conf/iccsa/NasirSJA25} used LLMs' hidden states as embeddings to fine-tune a smaller classifier. Nevertheless, none of the work explored massive tuning and generalization of LLMs to several hate speech tasks with different label schemes, hierarchies, and label types.






\section{Methodology}
We first describe our methodology for unifying datasets and converting them into the instructions, followed by our strategy for training on the unified dataset. 

\subsection{Data Collation}
We source several English datasets, with the aim of including a diverse set of label types and domains. The resulting compilation is a superset of over 600k examples, shown in Table \ref{tab:datasets}. Additionally, we also report all corresponding licenses of the datasets in Appendix~\ref{sec:app_licenses} with several datasets samples examples in Appendix~\ref{sec:app_datasets_examples}. We divide the task types into 4 distinct categories:

\begin{itemize}[noitemsep]
    \item Binary classification
    \item Multiclass classification
    \item Multilabel classification
    \item Generation
\end{itemize}

\noindent Datasets labeled as `hierarchical' include multiple tasks, where the main task is typically a binary classification, such as `hateful/not', and the subsequent tasks only apply to those labeled as `hateful'. 

Our test includes 9 of these datasets, which we selected primarily due to the reproducibility of their train/test split, allowing for fair comparison. The resulting test superset includes tasks from each of our 4 categories. 

\paragraph{Deduplication}
As many of the datasets are sourced from the same platforms, we deduplicate our training set to improve the quality. We first deduplicate based on the text/label pair, keeping only 1 copy. For examples with conflicting labels, we distinguish whether the examples are from the same dataset or not. If they are from the same dataset (as was largely the case with ToxiCR), we remove both copies. For cross-dataset duplicates, we temporarily binarize labels, and if they are still different, we keep the example with the `hateful or offensive' label. This only affected 75 (0.01\%) examples. In total, 3.73\% (25147 examples) of the dataset is removed via deduplication. The final size of the dataset after deduplication is 649761, as presented in Table \ref{tab:datasets}.
Full statistics of the appeared labels in the final aggregated dataset is presented in Appendix~\ref{app:labels_statistics}.

\begin{table*}[!htbp]\centering
\footnotesize
\begin{tabular}{lrrrrrrrr}\toprule
Dataset &Type &HIPPO &Indiv. &GPT5 &Best &Best Arch. &Best Source \\\midrule
AMI18 - A &Binary &\textbf{81.0} &74.8 &70.9 &71.0 &BERT &\citet{muti-barron-cedeno-2022-checkpoint} \\
AMI18 - B &Binary x2 &\textbf{73.4} &64.3 &44.4 &42.9 &BERT &\citet{pamungkas2020misogyny} \\
EDOS - A &Binary &83.7 &85.0 &73.0 &\textbf{88.4} &PaLM &\citet{sorensen-etal-2023-juage} \\
EDOS - B &Multiclass &71.2 &65.5 &41.9 &\textbf{73.3} &PaLM &\citet{sorensen-etal-2023-juage} \\
EDOS - C &Multiclass &51.8 &47.5 &41.0 &\textbf{56.1} &DeBERTa-v3 &\citet{zhou-2023-pinganlifeinsurance} \\
HASOC19 - A &Binary &76.7 &79.0 &72.7 &\textbf{79.5} &RoBERTa$^\dagger$ &\citet{kovacs2021challenges} \\
HASOC19 - B &Multiclass &\textbf{58.5} &56.7 &54.8 &54.5 &BERT &\citet{mishra20193idiots} \\
HASOC19 - C &Binary &52.8 &52.3 &\textbf{58.0} &51.1 &BERT &\citet{mishra20193idiots} \\
HatEval19 - A &Binary &56.8 &55.7 &66.8 &\textbf{70.8} &Flan-UL2 &\citet{zhang-etal-2024-sentiment} \\
HatEval19 - B &Binary x2 &\textbf{80.0} &77.8 &35.0 &62.0 &DistilBERT$^\dagger$ &\citet{atapattu2020automated} \\
HateXplain &Multiclass &69.5 &69.1 &40.8 &\textbf{69.9} &BERT &\citet{kim-etal-2022-hate} \\
Jigsaw &Multilabel &\textbf{80.7} &80.2 &61.5 &62.5 &BERT$^\dagger$ &\citet{mazari2024bert} \\
OffensEval20 - A &Binary &\textbf{92.8} &90.6 &89.7 &92.0 &RoBERTa$^\dagger$ &\citet{wiedemann-etal-2020-uhh} \\
OffensEval20 - B &Binary &66.2 &62.3 &67.1 &\textbf{74.6} &ERNIE 2.0$^\dagger$ &\citet{wang-etal-2020-galileo} \\
OffensEval20 - C &Multiclass &70.3 &67.0 &60.7 &\textbf{71.5} &ERNIE 2.0$^\dagger$ &\citet{wang-etal-2020-galileo} \\
ParaDetox &Generation &64.8 &65.2 &35.3 &\textbf{74.2} &DeepSeek &\citet{hosseinbeigi2025metadetox} \\ 
ToxicSpans &Generation &64.3 &63.1 &25.9 &\textbf{70.8} &BERT$^\dagger$ &\citet{zhu-etal-2021-hitsz} \\ \midrule
Average & &\textbf{70.3} &68.0 &55.3 &68.5 & & \\
\bottomrule
\end{tabular}

\caption{Comparison of our unified model (\hippo, based on \texttt{Qwen3-4B-Instruct-2507}) to individually trained models (Indiv.), GPT5-mini, and to the best performing models in literature. \textbf{Bold} values denote the best results per dataset. $\dagger$ denotes ensembles.}\label{tab:sota}
\end{table*}

\subsection{Instruction Construction}
Our general approach is to convert every individual dataset into a conversational format, with user prompts and assistant answers. Our user prompt depends on the task, but all start with the generic message, followed by the input text, followed by a task-specific message, as shown below for binary classification.
\begin{quote} \textbf{User:} You are an expert on hate speech, and you are tasked with answering questions about the following text. Respond only with the answer, no other text: \{\texttt{text}\} \\
Should the text be classified as \{\texttt{category}\}? Answer with `yes' or `no'.\\
\textbf{Assistant:} \{yes / no\}

\end{quote}
The \texttt{category} could be ``hateful'', for example. For datasets with a hierarchical structure, we pose the questions in a back-and-forth conversational manner. For example, continuing from the previous example (assuming the answer is `yes'):
\begin{quote}
\textbf{User:} Which category does the text belong to? Categories: \{\texttt{categories}\} \\
\textbf{Assistant:} \{\texttt{category\_choice}\}
\end{quote}
This prompt is used for multiclass tasks. For the multiclass and multilabel tasks, we also prepend each category with letters (`A. ', `B. ', `C. ', etc.) so that the label for these tasks is simply the letter (or letters for multilabel). We found that this increases performance substantially over generating the exact names of the classes. For the multilabel case, the label is a comma-separated list of the present labels. We also consistently order the labels alphabetically, so that the model does not receive conflicting signals during training.

For generative tasks, there is no shared prompt, as every generative dataset's task or tasks are unique. For example, for ToxicSpans, which we treat as a generative task, has the following prompt:
\begin{quote}
\textbf{User:} Identify the part(s) of the given text that is/are considered toxic. If there are multiple spans, separate them with semicolons. Write N/A if the text is not toxic. \\
\textbf{Assistant:} \{\texttt{span}\}
\end{quote}
Other datasets, such as DialoCONAN, are even more open-ended, as the goal is to model a hate speech expert having a counter-narrative with someone saying harmful things. The conversational format naturally suits this type of dataset:
\begin{quote}
\textbf{User:} 
\{\texttt{narrative\_1}\} \\
Have a dialogue with the author of the original text. Provide and maintain a counter-narrative throughout the dialogue. \\
\textbf{Assistant:} \{\texttt{counter\_narrative\_1}\} \\
\textbf{User:} \{\texttt{narrative\_2}\} \\
\textbf{Assistant:} \{\texttt{counter\_narrative\_2}\} \\
... \\
\textbf{User:} \{\texttt{narrative\_n}\} \\
\textbf{Assistant:} \{\texttt{counter\_narrative\_n}\} 
\end{quote}
We include an exhaustive list of our prompts in Appendix \ref{app:prompts}.

\subsection{Model Selection}
We experiment with various LLMs of 4 billion parameters, including Qwen3 \cite{qwen3technicalreport}, Llama3 \cite{dubey2024llama}, and Phi4 \cite{abdin2024phi}. We show the best-performing model---\textbf{\hippo}  based on \textbf{\texttt{Qwen3-4B-Instruct-2507}}, in Results (\S\ref{sect:results})---with the rest of LLMs in Appendix \ref{app:models}. We use a smaller model with 4 billion parameters for our main experiments to limit ecological impact as well as to demonstrate the utility of instruction tuning for hate speech already from this size of computationally-accessible models. Nevertheless, we also finetune a larger (32B) and even smaller (0.6B) model to gauge the effect of model size. 


\subsection{Training}
Training follows a typical instruction tuning setup. We apply QLoRA \cite{dettmers2023qlora} to enable efficient fine-tuning, and we train the model only on the outputs of the assistant. Training on an Nvidia H100 GPU, the 0.6B model takes around 10 hours, the 4B models take around 1 day, and the 32B model takes around 4 days. Additional training hyperparameters can be found in Appendix \ref{app:hyperparams}.

\section{Results} \label{sect:results}
We go through the results as they relate to the research questions one-by-one. The scores shown in the following tables and figures are macro-F1, with some exceptions:

\begin{itemize}[noitemsep]
    \item AMI18 task B and HatEval19 task B are both composed of 2 subtasks, with their F1 scores averaged. 
    \item ToxicSpans uses a character-based F1 score based on the overlap of the predicted span and label span.
    \item Jigsaw uses an average F1 score per-label. 
    \item ParaDetox uses the joint scoring method from \citet{dementieva2025overview} that combines style-transfer accuracy, content preservation, and fluency. 
\end{itemize}

More detailed results of \hippo with confusion matrices for the classification-based test datasets are presented in Appendix~\ref{app:extended_results}.

\subsection{Comparison to State-of-the-art}
\begin{quote}
\textit{\textbf{RQ1}: \rqone}
\end{quote}
Table \ref{tab:sota} shows our main results on our test superset. We compare the performance of our unified training to training on each individual training split of each test set, finding that indeed our unified training brings a benefit in 14 out of 17 cases. Moreover, our unified approach outperforms a strong closed-source model, GPT5-mini.

\begin{table}[!htbp]
\centering
\footnotesize
\begin{tabular}{lrrrrr}\toprule
Dataset &\wbox{0.6B} &\wbox{4B} &\wbox{32B} &\wbox{Best} \\\midrule
AMI18 - A &\wbox{28.1} &\bbox{81.0} &\bbox{\textbf{82.9}} &\wbox{71.0} \\
AMI18 - B &\wbox{26.0} &\bbox{73.4} &\bbox{\textbf{75.5}} &\wbox{42.9} \\
EDOS - A &\wbox{23.5} &\wbox{83.7} &\wbox{87.0} &\wbox{\textbf{88.4}} \\
EDOS - B &\wbox{42.0} &\wbox{71.2} &\bbox{\textbf{74.7}} &\wbox{73.3} \\
EDOS - C &\wbox{27.5} &\wbox{51.8} &\bbox{\textbf{57.9}} &\wbox{56.1} \\
HASOC19 - A &\wbox{22.2} &\wbox{76.7} &\wbox{76.6} &\wbox{\textbf{79.5}} \\
HASOC19 - B &\wbox{31.6} &\bbox{\textbf{58.5}} &\bbox{58.0} &\wbox{54.5} \\
HASOC19 - C &\wbox{47.6} &\bbox{52.8} &\bbox{\textbf{52.7}} &\wbox{51.1} \\
HatEval19 - A &\wbox{20.1} &\wbox{56.8} &\wbox{59.8} &\wbox{\textbf{70.8}} \\
HatEval19 - B &\wbox{41.0} &\bbox{\textbf{80.0}} &\bbox{79.8} &\wbox{62.0} \\
HateXplain &\wbox{16.1} &\wbox{69.5} &\bbox{\textbf{70.8}} &\wbox{69.9} \\
Jigsaw &\wbox{53.7} &\bbox{\textbf{80.7}} &\bbox{\textbf{80.7}} &\wbox{62.5} \\
OffensEval20 - A &\wbox{27.8} &\bbox{\textbf{92.8}} &\bbox{92.5} &\wbox{92.0} \\
OffensEval20 - B &\wbox{39.9} &\wbox{66.2} &\wbox{65.2} &\wbox{\textbf{74.6}} \\
OffensEval20 - C &\wbox{21.1} &\wbox{70.3} &\bbox{\textbf{75.4}} &\wbox{71.5} \\
ParaDetox &\wbox{10.3} &\wbox{64.8} &\wbox{65.0} &\wbox{\textbf{74.2}} \\ 
ToxicSpans &\wbox{21.9} &\wbox{64.3} &\wbox{64.2} &\wbox{\textbf{70.8}} \\
\midrule
Average &\wbox{29.4} &\bbox{70.3} &\bbox{\textbf{71.7}} &\wbox{68.5} \\
\bottomrule
\end{tabular}
\caption{Effectiveness of instruction tuning on varying model sizes, compared to the SOTA. \bbox{Blue} values indicate scores that outperform the SOTA. \textbf{Bold} values denote the best results per dataset.} \label{tab:model_size}
\end{table}


Compared to the best-known models in the literature, the unified training approach wins on 7 of 17 tasks and achieves higher average performance. Notably, many state-of-the-art systems rely on task-specific ensembling, whereas we use no task-specific methods beyond prompting. In addition, competing non-BERT models often have substantially larger parameter counts (i.e., Flan-UL2 at 20B, PaLM at 62B, and DeepSeek at 671B), while our model is based on Qwen3-4B. As shown in Appendix~\ref{app:extended_results}, our single model also generalizes across diverse tasks and label granularities, enabling a unified approach to hate speech mitigation.

In Table \ref{tab:model_size}, we show that a 32B parameter model has even better performance overall, winning in 11 of 17 cases versus the state-of-the-art. This indicates that the instruction tuning setup is even more effective at larger model sizes. At smaller model sizes, it appears ineffective. The 0.6B model performs particularly poorly, worse than random chance in many circumstances. This is because we do not constrain generation, so the model frequently does not output `yes' or `no' for binary classification tasks, for example.  

Overall, the results show that from a unified, single-step training of a modest-sized LLM of at least 4 billion parameters, one can achieve performance competitive with several highly-tuned, task-specific models. We continue the further comparison of our \hippo based on 4B-sized Qwen3.

\subsection{Prompting and Cross-task Transfer} \label{subsect:xtask}
\begin{quote}
\textit{\textbf{RQ2}: \rqtwo}    
\end{quote}
In Table \ref{tabl:xtask}, we first compare our fully-trained 4B model (i.e., \hippo  in Table \ref{tab:sota}) to zero-shot prompting the same base model. Our zero-shot prompts are identical to the instructions used in training. We additionally post-processed the zero-shot outputs to ensure variations of the correct answer (e.g., `yes' versus `Yes.') do not negatively affect the results.

We do not include results for few-shot prompting, as we were unable to formulate a few-shot prompt that outperformed, or even matched performance of the zero-shot prompt. This has been previously observed in other works on prompting for hate speech detection, where few-shot prompting performed worse than zero-shot \cite{guo2023investigation}, or its benefits were inconsistent across languages \cite{ghorbanpour-etal-2025-prompting}.

Nevertheless, we see that the fine-tuned model clearly outperforms our zero-shot prompt. While the prompts used are likely suboptimal, finding the optimal prompt for each task is time-consuming, likely more so than simply instruction tuning the model. 

Concerning the leave-one-out (LOO) column in Table \ref{tabl:xtask}, we turn to our third research question:
\begin{quote}
\textit{\textbf{RQ3}: \rqthree}
    
\end{quote}

To answer this, we train 9 additional models for each test dataset, each with the dataset's corresponding training split left out. 

\begin{table}[!tp]
\centering
\footnotesize

\begin{tabular}{lrrrr}\toprule
Dataset &\wbox{Zero-shot} &\wbox{LOO} &\wbox{Full} \\\midrule
AMI18 - A &\wbox{59.8} &\bbox{\textbf{78.4}} &\bbox{\textbf{81.0}} \\
AMI18 - B$^\dagger$ &\wbox{41.6} &\bbox{42.1} &\bbox{\textbf{73.4}} \\
EDOS - A &\wbox{66.7} &\bbox{76.0} &\bbox{\textbf{83.7}} \\
EDOS - B$^\ddagger$ &\wbox{37.9} &\bbox{43.4} &\bbox{\textbf{71.2}} \\
EDOS - C$^\ddagger$ &\wbox{20.2} &\bbox{27.6} &\bbox{\textbf{51.8}} \\
HASOC19 - A &\wbox{71.5} &\wbox{70.3} &\bbox{\textbf{76.7}} \\
HASOC19 - B &\wbox{31.2} &\bbox{52.0} &\bbox{\textbf{58.5}} \\
HASOC19 - C &\wbox{\textbf{63.4}} &\wbox{44.0} &\wbox{52.8} \\
HatEval19 - A &\wbox{\textbf{58.1}} &\wbox{52.5} &\wbox{56.8} \\
HatEval19 - B$^\dagger$ &\wbox{33.3} &\wbox{29.3} &\bbox{\textbf{80.0}} \\
HateXplain &\wbox{46.8} &\bbox{60.1} &\bbox{\textbf{69.5}} \\
Jigsaw$^\ddagger$ &\wbox{59.2} &\bbox{70.1} &\bbox{\textbf{80.7}} \\
OffensEval20 - A &\wbox{80.8} &\bbox{88.8} &\bbox{\textbf{92.8}} \\
OffensEval20 - B &\wbox{\textbf{72.8}} &\wbox{66.2} &\wbox{66.2} \\
OffensEval20 - C &\wbox{43.0} &\bbox{69.8} &\bbox{\textbf{70.3}} \\
ParaDetox$^\ddagger$ &\wbox{48.0} &\wbox{46.4} &\bbox{\textbf{64.8}} \\ 
ToxicSpans &\wbox{22.1} &\bbox{42.3} &\bbox{\textbf{64.3}} \\
\midrule
Average &\wbox{50.4} &\bbox{56.4} &\bbox{\textbf{70.3}} \\
\bottomrule
\end{tabular}

\caption{Comparison of zero-shot to leave-one-out (LOO) and full \hippo (i.e., unified) training. $\ddagger$ denotes tasks that do not have equivalent tasks in training of \texttt{Qwen3-4B}. $\dagger$ denotes partial equivalence: AMI18-B and HatEval19-B are composed of 2 tasks, both of which include 1 task that has no similar task in training. \bbox{Blue} values indicate scores that outperform the zero-shot baseline. \textbf{Bold} values denote the best results per dataset.} \label{tabl:xtask}
\end{table}

As we can see, the results are generally better than the zero-shot performance, showing a clear cross-task learning capability. Of the 17 tasks, only 6 have a decreased performance: HatEval19-A and B, HASOC19-A and C, OffensEval20-B, and ParaDetox. 

\begin{figure}[tp !]
    \centering
    \includegraphics[width=\linewidth]{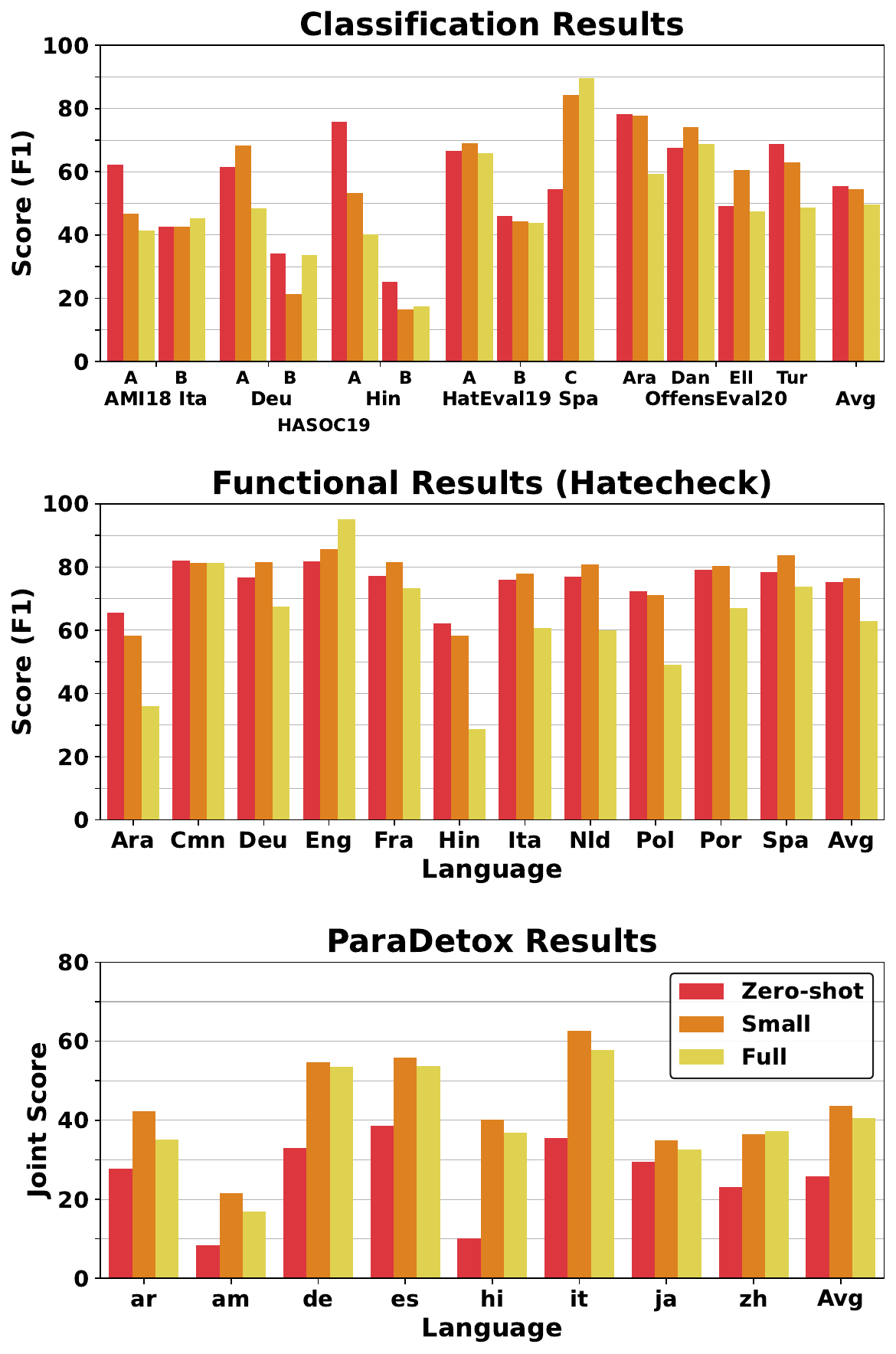}
    \caption{Cross-lingual and functional performance of the zero-shot, minimally-trained, and fully-trained \hippo. Note: Unlike for English, HASOC19 does not have a Task C for German and Hindi.}
    \label{fig:xling}
\end{figure}

Of the 6 tasks, HatEval19-B, HASOC19-C, and OffensEval20-B  all have the same goal of determining whether a text is targeted or not. The issue may be due to dissimilar annotation guidelines or collection methods, as HatEval19 has a near-balanced training set (51\% targeted, 49\% untargeted), while HASOC19 and OffensEval20 have very lopsided distributions (85/15 and 88/12, respectively).

\begin{figure}
    \centering
    \includegraphics[width=\linewidth]{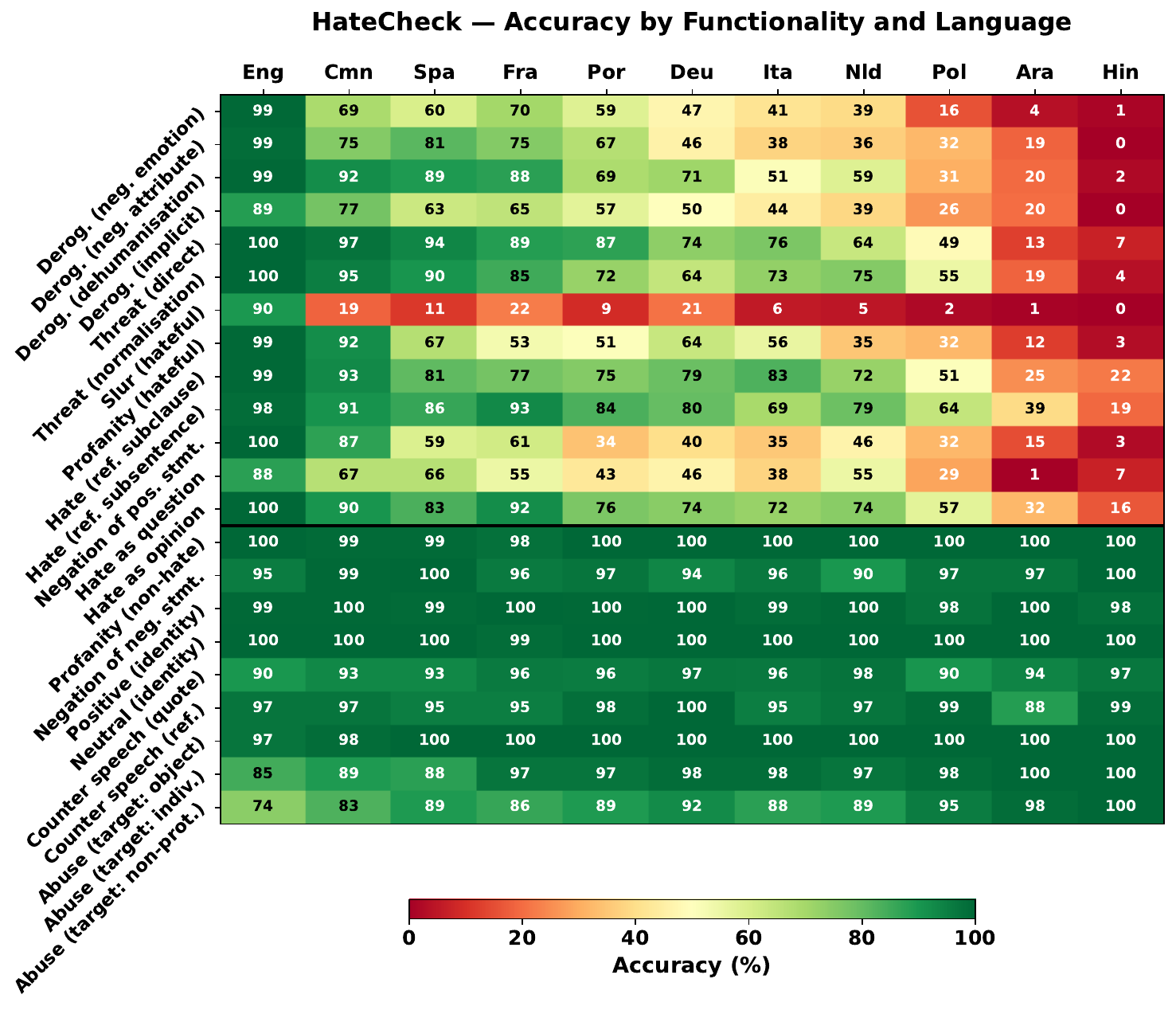}
    \caption{Performance of the fully-trained HIPPO on all languages (ordered by performance) of HateCheck, per functional task. Above the black line are hateful functionalities, below are non-hateful. Only functionalities shared across all languages are shown.}
    \label{fig:xling_degrade}
\end{figure}

\subsection{Cross-lingual Transfer}
\begin{quote}
\textit{\textbf{RQ4}: \rqfour}
\end{quote}
While we focus on English for this work, high-quality hate speech processing is necessary for most languages. And since our base Qwen model has been pretrained multilingually, we therefore test whether our finetuning has damaged or helped its non-English hate speech processing. We compare our zero-shot model with a fully-finetuned model, as well as an additional model trained on only 1k examples from each dataset. This smaller training set is intended to expose the model to the domain while not overfitting to English. 
We evaluate along 3 separate axes: classification, generation, and functional performance. 
For testing, we keep the prompts in English and only swap out the text and additional task information, such as class names. We show results in Figure \ref{fig:xling}.


In general, we observe that the fully trained model’s performance drops on both classification and functional tasks compared to zero-shot. As shown in Figure \ref{fig:xling_degrade}, this degradation varies by functionality in HateCheck. Arabic and Hindi perform worst, likely due to their distinct scripts, while Chinese performs better, possibly reflecting stronger representation in pretraining data. Performance on slurs is notably poor, consistent with their language-specific nature. In contrast, performance on non-hateful content remains nearly perfect across languages, suggesting a greater familiarity with non-toxic text from pretraining.


Returning to Figure \ref{fig:xling}, ParaDetox performance improves over zero-shot, which is unexpected. However, qualitative analysis shows the zero-shot model often translates inputs into English despite no instruction to do so, lowering its scores.

For the minimally trained model, performance is competitive on classification tasks, generally higher on HateCheck, and substantially better on ParaDetox. This suggests that even limited English-only instruction tuning can yield positive cross-lingual transfer across diverse tasks.

There are many non-English hate speech training datasets available, so for truly better multilingual performance, we expect it would be highly beneficial to incorporate those into training. We leave this for future work. At the same time, we can see how the model is capable not only of classification, but for generation tasks already for multiple languages, which can serve as a strong out-of-the-box baseline.

\subsection{Further Discussion}
We additionally discuss two topics which may have a considerable impact on our models' performances, label disagreement and data contamination, in Appendix \ref{app:discussion}.


\section{Conclusion}

As hate speech can rapidly evolve in order to evade content moderation, the ability to perform generalizable hate speech mitigation has become increasingly important for practitioners who wish to curb the use of hate speech. To advance this goal, we introduced a unified instruction-style carefully curated training corpus constructed from 36 heterogeneous datasets, incorporating a wide range of harmful content classification tasks as well as several proactive speech generation tasks. We further evaluated multiple LLM families and sizes, and presented \hippo, a model based on the Qwen3-4B variant, as an effective generalist solution.

We demonstrated that our generalist model performs at least on par with or significantly outperforms prior task-specific state-of-the-art BERT-style models, as well as prompt-based LLM approaches. We also saw a distinct increase in performance of our model on held-out sets, indicating an ability to generalize to new tasks which is quite impossible for BERT-based specialists. Finally, we showed that this generalization also extends to some multilingual tasks, although further training on multilingual data would likely benefit non-English languages further. We believe that our findings show the effectiveness of obtaining LLMs-based generalists for more robust harmful speech detection and proactive moderation across various domains.

\section*{Limitations}
Our work is limited to training on only English datasets. While we do not expect that the trends would be different in other languages, we cannot know for sure without further experimentation. 
Training on multiple languages at once is also something we do not test, and could be more promising for better cross-lingual performance. However, it is also possible that cultural differences result in conflicting annotation styles that hurt performance.

We also do not test on a large number of prompting styles. For the finetuned model, the prompting style is likely not very impactful due to the additional training. For the zero-shot model, it is much more impactful. Our zero-shot results should not be seen as optimal zero-shot performance, but rather as performance on a manually crafted yet unoptimized prompt. There is additional information that could improve a prompt, such as definitions or annotation guidelines. Due to the lack of availability of this information for many datasets, we do not include it in our prompt.

We additionally do not incorporate or test model reasoning for hate speech tasks. Our GPT5-mini results used the default reasoning effort, ``medium'', however we did not test other effort levels. Reasoning has been shown to be beneficial for many tasks, however we are not aware of any hate speech data with annotated reasoning, and the quality of LLMs' hate speech detection is not good enough to make use of synthetic reasoning data.  

\section*{Ethics Statement}
Automatic moderation has become a promising application of modern NLP-based models, including recent LLMs. While such technologies may raise concerns about censorship, our work is strictly positioned as an investigation into the capabilities and limitations of modern language technologies, with the explicit goal of improving user safety in online environments. For this reason, we release our code and models under the OpenRAIL-S license, which restricts AI-based solutions usage to responsible and socially beneficial applications. The licenses of all used (and, later, our published) resources can be found in Appendix~\ref{sec:app_licenses}.

Nevertheless, there remains a risk of misuse for dangerous scenarios---aggressive censorship, jail-breaking of models for hate behavior, or even generation of more hateful content. The generation of hateful or toxic language can already be achieved through simpler ways---for example, by injecting profane expressions or by relying on existing text style transfer models. Despite this accessibility, there has been no evidence of a large-scale surge in synthetic hateful content. Companies and researchers instantly continuing to improve safe mechanisms in LLMs making it incredibly difficult to jail-break with simple toxic injections.

In addition, we emphasize that effective content moderation cannot rely solely on automated systems. Responsible deployment requires sustained user-centered research and careful consideration of social context. Decisions about moderation policies should remain with the communities themselves, ensuring that safety mechanisms reflect their values, norms, and expectations. 

At the same time, our proposed generalist hate speech mitigation model can now enable platform moderators to define more flexible and fine-grained content moderation guidelines, supporting scalable and consistent hate speech detection across large and diverse online environments. Then, our model outputs should serve as a recommendation for further moderation decisions, not as the ultimate final answer. We truly believe that our findings will serve as a base for future robust harmful human-written or AI-generated content detection for a safe online environment.


\bibliography{custom}

\begin{thebibliography}{90}
\providecommand{\natexlab}[1]{#1}

\bibitem[{Abdin et~al.(2024)Abdin, Aneja, Behl, Bubeck, Eldan, Gunasekar,
  Harrison, Hewett, Javaheripi, Kauffmann et~al.}]{abdin2024phi}
Marah Abdin, Jyoti Aneja, Harkirat Behl, S{\'e}bastien Bubeck, Ronen Eldan,
  Suriya Gunasekar, Michael Harrison, Russell~J Hewett, Mojan Javaheripi, Piero
  Kauffmann, and 1 others. 2024.
\newblock Phi-4 technical report.
\newblock \emph{arXiv preprint arXiv:2412.08905}.

\bibitem[{Antypas and
  Camacho-Collados(2023)}]{antypas-camacho-collados-2023-robust}
Dimosthenis Antypas and Jose Camacho-Collados. 2023.
\newblock \href {https://doi.org/10.18653/v1/2023.woah-1.25} {Robust hate
  speech detection in social media: A cross-dataset empirical evaluation}.
\newblock In \emph{The 7th Workshop on Online Abuse and Harms (WOAH)}, pages
  231--242, Toronto, Canada. Association for Computational Linguistics.

\bibitem[{Arkhoshghalb(2019)}]{toosi2019twitterhate}
Armin Arkhoshghalb. 2019.
\newblock Twitter sentiment analysis: Hatred speech.
\newblock
  \url{https://www.kaggle.com/datasets/arkhoshghalb/twitter-sentiment-analysis-hatred-speech/data}.
\newblock Kaggle Dataset.

\bibitem[{Arora et~al.(2024)Arora, Nakov, Hardalov, Sarwar, Nayak, Dinkov,
  Zlatkova, Dent, Bhatawdekar, Bouchard, and
  Augenstein}]{DBLP:journals/csur/AroraNHSNDZDBBA24}
Arnav Arora, Preslav Nakov, Momchil Hardalov, Sheikh~Muhammad Sarwar, Vibha
  Nayak, Yoan Dinkov, Dimitrina Zlatkova, Kyle Dent, Ameya Bhatawdekar,
  Guillaume Bouchard, and Isabelle Augenstein. 2024.
\newblock \href {https://doi.org/10.1145/3603399} {Detecting harmful content on
  online platforms: What platforms need vs. where research efforts go}.
\newblock \emph{{ACM} Comput. Surv.}, 56(3):72:1--72:17.

\bibitem[{Atapattu et~al.(2020)Atapattu, Herath, Zhang, and
  Falkner}]{atapattu2020automated}
Thushari Atapattu, Mahen Herath, Georgia Zhang, and Katrina Falkner. 2020.
\newblock Automated detection of cyberbullying against women and immigrants and
  cross-domain adaptability.
\newblock \emph{arXiv preprint arXiv:2012.02565}.

\bibitem[{Atwell et~al.(2022)Atwell, Hassan, and
  Alikhani}]{atwell-etal-2022-appdia}
Katherine Atwell, Sabit Hassan, and Malihe Alikhani. 2022.
\newblock \href {https://aclanthology.org/2022.coling-1.530} {{APPDIA}: A
  discourse-aware transformer-based style transfer model for offensive social
  media conversations}.
\newblock In \emph{Proceedings of the 29th International Conference on
  Computational Linguistics}, pages 6063--6074, Gyeongju, Republic of Korea.
  International Committee on Computational Linguistics.

\bibitem[{Basile et~al.(2019)Basile, Bosco, Fersini, Nozza, Patti, Pardo,
  Rosso, and Sanguinetti}]{basile2019semeval}
Valerio Basile, Cristina Bosco, Elisabetta Fersini, Debora Nozza, Viviana
  Patti, Francisco Manuel~Rangel Pardo, Paolo Rosso, and Manuela Sanguinetti.
  2019.
\newblock Semeval-2019 task 5: Multilingual detection of hate speech against
  immigrants and women in twitter.
\newblock In \emph{Proceedings of the 13th international workshop on semantic
  evaluation}, pages 54--63.

\bibitem[{Bonaldi et~al.(2022)Bonaldi, Dellantonio, Tekiro{\u{g}}lu, and
  Guerini}]{bonaldi-etal-2022-human}
Helena Bonaldi, Sara Dellantonio, Serra~Sinem Tekiro{\u{g}}lu, and Marco
  Guerini. 2022.
\newblock \href {https://doi.org/10.18653/v1/2022.emnlp-main.549}
  {Human-machine collaboration approaches to build a dialogue dataset for hate
  speech countering}.
\newblock In \emph{Proceedings of the 2022 Conference on Empirical Methods in
  Natural Language Processing}, pages 8031--8049, Abu Dhabi, United Arab
  Emirates. Association for Computational Linguistics.

\bibitem[{Caselli et~al.(2021)Caselli, Basile, Mitrovi{\'c}, and
  Granitzer}]{caselli-etal-2021-hatebert}
Tommaso Caselli, Valerio Basile, Jelena Mitrovi{\'c}, and Michael Granitzer.
  2021.
\newblock \href {https://doi.org/10.18653/v1/2021.woah-1.3} {{H}ate{BERT}:
  Retraining {BERT} for abusive language detection in {E}nglish}.
\newblock In \emph{Proceedings of the 5th Workshop on Online Abuse and Harms
  (WOAH 2021)}, pages 17--25, Online. Association for Computational
  Linguistics.

\bibitem[{Cercas~Curry et~al.(2021)Cercas~Curry, Abercrombie, and
  Rieser}]{cercas-curry-etal-2021-convabuse}
Amanda Cercas~Curry, Gavin Abercrombie, and Verena Rieser. 2021.
\newblock \href {https://doi.org/10.18653/v1/2021.emnlp-main.587}
  {{C}onv{A}buse: Data, analysis, and benchmarks for nuanced abuse detection in
  conversational {AI}}.
\newblock In \emph{Proceedings of the 2021 Conference on Empirical Methods in
  Natural Language Processing}, pages 7388--7403, Online and Punta Cana,
  Dominican Republic. Association for Computational Linguistics.

\bibitem[{Chung et~al.(2019)Chung, Kuzmenko, Tekiroglu, and
  Guerini}]{chung-etal-2019-conan}
Yi-Ling Chung, Elizaveta Kuzmenko, Serra~Sinem Tekiroglu, and Marco Guerini.
  2019.
\newblock \href {https://doi.org/10.18653/v1/P19-1271} {{CONAN} - {CO}unter
  {NA}rratives through nichesourcing: a multilingual dataset of responses to
  fight online hate speech}.
\newblock In \emph{Proceedings of the 57th Annual Meeting of the Association
  for Computational Linguistics}, pages 2819--2829, Florence, Italy.
  Association for Computational Linguistics.

\bibitem[{Cjadams et~al.(2017)Cjadams, Sorensen, Elliott, Dixon, McDonald,
  nithum, and Cukierski}]{jigsaw-toxic-comment-classification-challenge}
Cjadams, Jeffrey Sorensen, Julia Elliott, Lucas Dixon, Mark McDonald, nithum,
  and Will Cukierski. 2017.
\newblock Toxic comment classification challenge.
\newblock
  \href{https://kaggle.com/competitions/jigsaw-toxic-comment-classification-challenge}{https://kaggle.com/competitions/jigsaw-toxic-comment-classification-challenge}.
\newblock Kaggle.

\bibitem[{Corazza et~al.(2019)Corazza, Menini, Cabrio, Tonelli, and
  Villata}]{DBLP:conf/clic-it/CorazzaMCTV19}
Michele Corazza, Stefano Menini, Elena Cabrio, Sara Tonelli, and Serena
  Villata. 2019.
\newblock \href {https://ceur-ws.org/Vol-2481/paper22.pdf} {Cross-platform
  evaluation for italian hate speech detection}.
\newblock In \emph{Proceedings of the Sixth Italian Conference on Computational
  Linguistics, Bari, Italy, November 13-15, 2019}, volume 2481 of \emph{{CEUR}
  Workshop Proceedings}. CEUR-WS.org.

\bibitem[{Costa{-}juss{\`{a}} et~al.(2022)Costa{-}juss{\`{a}}, Cross,
  {\c{C}}elebi, Elbayad, Heafield, Heffernan, Kalbassi, Lam, Licht, Maillard,
  Sun, Wang, Wenzek, Youngblood, Akula, Barrault, Gonzalez, Hansanti, Hoffman,
  Jarrett, Sadagopan, Rowe, Spruit, Tran, Andrews, Ayan, Bhosale, Edunov, Fan,
  Gao, Goswami, Guzm{\'{a}}n, Koehn, Mourachko, Ropers, Saleem, Schwenk, and
  Wang}]{costa2022no}
Marta~R. Costa{-}juss{\`{a}}, James Cross, Onur {\c{C}}elebi, Maha Elbayad,
  Kenneth Heafield, Kevin Heffernan, Elahe Kalbassi, Janice Lam, Daniel Licht,
  Jean Maillard, Anna~Y. Sun, Skyler Wang, Guillaume Wenzek, Al~Youngblood,
  Bapi Akula, Lo{\"{\i}}c Barrault, Gabriel~Mejia Gonzalez, Prangthip Hansanti,
  John Hoffman, and 19 others. 2022.
\newblock \href {https://doi.org/10.48550/ARXIV.2207.04672} {No language left
  behind: Scaling human-centered machine translation}.
\newblock \emph{CoRR}, abs/2207.04672.

\bibitem[{Davidson et~al.(2017)Davidson, Warmsley, Macy, and
  Weber}]{davidson2017automated}
Thomas Davidson, Dana Warmsley, Michael Macy, and Ingmar Weber. 2017.
\newblock Automated hate speech detection and the problem of offensive
  language.
\newblock In \emph{Proceedings of the international AAAI conference on web and
  social media}, volume~11, pages 512--515.

\bibitem[{De~Gibert et~al.(2018)De~Gibert, Perez, Garc{\'\i}a-Pablos, and
  Cuadros}]{de2018hate}
Ona De~Gibert, Naiara Perez, Aitor Garc{\'\i}a-Pablos, and Montse Cuadros.
  2018.
\newblock Hate speech dataset from a white supremacy forum.
\newblock \emph{arXiv preprint arXiv:1809.04444}.

\bibitem[{Dementieva et~al.(2025{\natexlab{a}})Dementieva, Babakov, Ronen,
  Ayele, Rizwan, Schneider, Wang, Yimam, Moskovskiy, Stakovskii, Kaufman,
  Elnagar, Mukherjee, and Panchenko}]{dementieva-etal-2025-multilingual}
Daryna Dementieva, Nikolay Babakov, Amit Ronen, Abinew~Ali Ayele, Naquee
  Rizwan, Florian Schneider, Xintong Wang, Seid~Muhie Yimam, Daniil Moskovskiy,
  Elisei Stakovskii, Eran Kaufman, Ashraf Elnagar, Animesh Mukherjee, and
  Alexander Panchenko. 2025{\natexlab{a}}.
\newblock \href {https://aclanthology.org/2025.coling-main.535} {Multilingual
  and explainable text detoxification with parallel corpora}.
\newblock In \emph{Proceedings of the 31st International Conference on
  Computational Linguistics}, pages 7998--8025, Abu Dhabi, UAE. Association for
  Computational Linguistics.

\bibitem[{Dementieva et~al.(2025{\natexlab{b}})Dementieva, Protasov, Babakov,
  Rizwan, Alimova, Brune, Konovalov, Muti, Liebeskind, Litvak
  et~al.}]{dementieva2025overview}
Daryna Dementieva, Vitaly Protasov, Nikolay Babakov, Naquee Rizwan, Ilseyar
  Alimova, Caroline Brune, Vasily Konovalov, Arianna Muti, Chaya Liebeskind,
  Marina Litvak, and 1 others. 2025{\natexlab{b}}.
\newblock Overview of the multilingual text detoxification task at pan 2025.
\newblock \emph{Working Notes of CLEF}.

\bibitem[{Dettmers et~al.(2023)Dettmers, Pagnoni, Holtzman, and
  Zettlemoyer}]{dettmers2023qlora}
Tim Dettmers, Artidoro Pagnoni, Ari Holtzman, and Luke Zettlemoyer. 2023.
\newblock Qlora: Efficient finetuning of quantized llms.
\newblock \emph{Advances in neural information processing systems},
  36:10088--10115.

\bibitem[{Devlin et~al.(2019)Devlin, Chang, Lee, and
  Toutanova}]{devlin-etal-2019-bert}
Jacob Devlin, Ming-Wei Chang, Kenton Lee, and Kristina Toutanova. 2019.
\newblock \href {https://doi.org/10.18653/v1/N19-1423} {{BERT}: Pre-training of
  deep bidirectional transformers for language understanding}.
\newblock In \emph{Proceedings of the 2019 Conference of the North {A}merican
  Chapter of the Association for Computational Linguistics: Human Language
  Technologies, Volume 1 (Long and Short Papers)}, pages 4171--4186,
  Minneapolis, Minnesota. Association for Computational Linguistics.

\bibitem[{D{\"o}nmez et~al.(2024)D{\"o}nmez, Vu, and
  Falenska}]{donmez-etal-2024-please}
Esra D{\"o}nmez, Thang Vu, and Agnieszka Falenska. 2024.
\newblock \href {https://doi.org/10.18653/v1/2024.emnlp-main.1019} {Please note
  that {I}{'}m just an {AI}: Analysis of behavior patterns of {LLM}s in
  (non-)offensive speech identification}.
\newblock In \emph{Proceedings of the 2024 Conference on Empirical Methods in
  Natural Language Processing}, pages 18340--18357, Miami, Florida, USA.
  Association for Computational Linguistics.

\bibitem[{Dubey et~al.(2024)Dubey, Jauhri, Pandey, Kadian, Al-Dahle, Letman,
  Mathur, Schelten, Yang, Fan et~al.}]{dubey2024llama}
Abhimanyu Dubey, Abhinav Jauhri, Abhinav Pandey, Abhishek Kadian, Ahmad
  Al-Dahle, Aiesha Letman, Akhil Mathur, Alan Schelten, Amy Yang, Angela Fan,
  and 1 others. 2024.
\newblock The llama 3 herd of models.
\newblock \emph{arXiv preprint arXiv:2407.21783}.

\bibitem[{ElSherief et~al.(2021)ElSherief, Ziems, Muchlinski, Anupindi,
  Seybolt, De~Choudhury, and Yang}]{elsherief-etal-2021-latent}
Mai ElSherief, Caleb Ziems, David Muchlinski, Vaishnavi Anupindi, Jordyn
  Seybolt, Munmun De~Choudhury, and Diyi Yang. 2021.
\newblock \href {https://doi.org/10.18653/v1/2021.emnlp-main.29} {Latent
  hatred: A benchmark for understanding implicit hate speech}.
\newblock In \emph{Proceedings of the 2021 Conference on Empirical Methods in
  Natural Language Processing}, pages 345--363, Online and Punta Cana,
  Dominican Republic. Association for Computational Linguistics.

\bibitem[{Fanton et~al.(2021)Fanton, Bonaldi, Tekiroğlu, and
  Guerini}]{fanton-2021-human}
Margherita Fanton, Helena Bonaldi, Serra~Sinem Tekiroğlu, and Marco Guerini.
  2021.
\newblock {Human-in-the-Loop for Data Collection: a Multi-Target Counter
  Narrative Dataset to Fight Online Hate Speech}.
\newblock In \emph{Proceedings of the 59th Annual Meeting of the Association
  for Computational Linguistics}. Association for Computational Linguistics.

\bibitem[{Fersini et~al.(2018)Fersini, Nozza, Rosso
  et~al.}]{fersini2018overview}
Elisabetta Fersini, Debora Nozza, Paolo Rosso, and 1 others. 2018.
\newblock Overview of the evalita 2018 task on automatic misogyny
  identification (ami).
\newblock In \emph{CEUR workshop proceedings}, volume 2263, pages 1--9.
  CEUR-WS.

\bibitem[{Fortuna et~al.(2019)Fortuna, Rocha~da Silva, Soler-Company, Wanner,
  and Nunes}]{fortuna-etal-2019-hierarchically}
Paula Fortuna, Jo{\~a}o Rocha~da Silva, Juan Soler-Company, Leo Wanner, and
  S{\'e}rgio Nunes. 2019.
\newblock \href {https://doi.org/10.18653/v1/W19-3510} {A
  hierarchically-labeled {P}ortuguese hate speech dataset}.
\newblock In \emph{Proceedings of the Third Workshop on Abusive Language
  Online}, pages 94--104, Florence, Italy. Association for Computational
  Linguistics.

\bibitem[{Fortuna et~al.(2020)Fortuna, Soler, and Wanner}]{fortuna2020toxic}
Paula Fortuna, Juan Soler, and Leo Wanner. 2020.
\newblock Toxic, hateful, offensive or abusive? what are we really classifying?
  an empirical analysis of hate speech datasets.
\newblock In \emph{Proceedings of the Twelfth Language Resources and Evaluation
  Conference}, pages 6786--6794.

\bibitem[{Founta et~al.(2018)Founta, Djouvas, Chatzakou, Leontiadis, Blackburn,
  Stringhini, Vakali, Sirivianos, and Kourtellis}]{founta2018large}
Antigoni Founta, Constantinos Djouvas, Despoina Chatzakou, Ilias Leontiadis,
  Jeremy Blackburn, Gianluca Stringhini, Athena Vakali, Michael Sirivianos, and
  Nicolas Kourtellis. 2018.
\newblock Large scale crowdsourcing and characterization of twitter abusive
  behavior.
\newblock In \emph{Proceedings of the international AAAI conference on web and
  social media}, volume~12.

\bibitem[{Gao and Huang(2017)}]{gao2017detecting}
Lei Gao and Ruihong Huang. 2017.
\newblock Detecting online hate speech using context aware models.
\newblock \emph{arXiv preprint arXiv:1710.07395}.

\bibitem[{Ghorbanpour et~al.(2025)Ghorbanpour, Dementieva, and
  Fraser}]{ghorbanpour-etal-2025-prompting}
Faeze Ghorbanpour, Daryna Dementieva, and Alexander Fraser. 2025.
\newblock \href {https://aclanthology.org/2025.woah-1.39} {Can prompting {LLM}s
  unlock hate speech detection across languages? a zero-shot and few-shot
  study}.
\newblock In \emph{Proceedings of the The 9th Workshop on Online Abuse and
  Harms (WOAH)}, pages 413--425, Vienna, Austria. Association for Computational
  Linguistics.

\bibitem[{Grimminger and Klinger(2021)}]{grimminger-klinger-2021-hate}
Lara Grimminger and Roman Klinger. 2021.
\newblock \href {https://aclanthology.org/2021.wassa-1.18} {Hate towards the
  political opponent: A {T}witter corpus study of the 2020 {US} elections on
  the basis of offensive speech and stance detection}.
\newblock In \emph{Proceedings of the Eleventh Workshop on Computational
  Approaches to Subjectivity, Sentiment and Social Media Analysis}, pages
  171--180, Online. Association for Computational Linguistics.

\bibitem[{Guo et~al.(2023)Guo, Hu, Mu, Shi, Zhao, Vishwamitra, and
  Hu}]{guo2023investigation}
Keyan Guo, Alexander Hu, Jaden Mu, Ziheng Shi, Ziming Zhao, Nishant
  Vishwamitra, and Hongxin Hu. 2023.
\newblock An investigation of large language models for real-world hate speech
  detection.
\newblock In \emph{2023 International Conference on Machine Learning and
  Applications (ICMLA)}, pages 1568--1573. IEEE.

\bibitem[{Hosseinbeigi et~al.(2025)Hosseinbeigi, Kelishami, Gheysari, and
  Rahimzadeh}]{hosseinbeigi2025metadetox}
Sara~Bourbour Hosseinbeigi, Amin~Saeidi Kelishami, Maryam Gheysari, and Fatemeh
  Rahimzadeh. 2025.
\newblock Metadetox at textdetox clef 2025: Detoxification with few-chain
  prompting.

\bibitem[{Huang et~al.(2023)Huang, Kwak, and An}]{huang2023chatgpt}
Fan Huang, Haewoon Kwak, and Jisun An. 2023.
\newblock \href {https://dl.acm.org/doi/10.1145/3543873.3587368} {Is chatgpt
  better than human annotators? potential and limitations of chatgpt in
  explaining implicit hate speech}.
\newblock In \emph{Companion proceedings of the ACM web conference 2023}, pages
  294--297.

\bibitem[{Huang(2025)}]{DBLP:journals/air/Huang25}
Tao Huang. 2025.
\newblock \href {https://doi.org/10.1007/S10462-025-11328-1} {Content
  moderation by {LLM:} from accuracy to legitimacy}.
\newblock \emph{Artif. Intell. Rev.}, 58(10):320.

\bibitem[{Jaki and Smedt(2019)}]{DBLP:journals/corr/abs-1910-07518}
Sylvia Jaki and Tom~De Smedt. 2019.
\newblock \href {https://arxiv.org/abs/1910.07518} {Right-wing german hate
  speech on twitter: Analysis and automatic detection}.
\newblock \emph{CoRR}, abs/1910.07518.

\bibitem[{Jiang et~al.(2024)Jiang, Liu, Zhong, Schaeffer, Ouyang, Han, and
  Koyejo}]{jiang2024investigating}
Minhao Jiang, Ken~Ziyu Liu, Ming Zhong, Rylan Schaeffer, Siru Ouyang, Jiawei
  Han, and Sanmi Koyejo. 2024.
\newblock Investigating data contamination for pre-training language models.
\newblock \emph{arXiv preprint arXiv:2401.06059}.

\bibitem[{Kennedy et~al.(2022)Kennedy, Atari, Davani, Yeh, Omrani, Kim,
  Coombs~Jr, Havaldar, Portillo-Wightman, Gonzalez
  et~al.}]{kennedy2022introducing}
Brendan Kennedy, Mohammad Atari, Aida~Mostafazadeh Davani, Leigh Yeh, Ali
  Omrani, Yehsong Kim, Kris Coombs~Jr, Shreya Havaldar, Gwenyth
  Portillo-Wightman, Elaine Gonzalez, and 1 others. 2022.
\newblock Introducing the gab hate corpus: defining and applying hate-based
  rhetoric to social media posts at scale.
\newblock \emph{Language Resources and Evaluation}, 56(1):79--108.

\bibitem[{Kennedy et~al.(2020)Kennedy, Bacon, Sahn, and von
  Vacano}]{kennedy2020constructing}
Chris~J Kennedy, Geoff Bacon, Alexander Sahn, and Claudia von Vacano. 2020.
\newblock Constructing interval variables via faceted rasch measurement and
  multitask deep learning: a hate speech application.
\newblock \emph{arXiv preprint arXiv:2009.10277}.

\bibitem[{Kim et~al.(2022)Kim, Lee, and Sohn}]{kim-etal-2022-hate}
Jiyun Kim, Byounghan Lee, and Kyung-Ah Sohn. 2022.
\newblock \href {https://aclanthology.org/2022.coling-1.577} {Why is it hate
  speech? masked rationale prediction for explainable hate speech detection}.
\newblock In \emph{Proceedings of the 29th International Conference on
  Computational Linguistics}, pages 6644--6655, Gyeongju, Republic of Korea.
  International Committee on Computational Linguistics.

\bibitem[{Kirk et~al.(2023)Kirk, Yin, Vidgen, and Röttger}]{kirkSemEval2023}
Hannah~Rose Kirk, Wenjie Yin, Bertie Vidgen, and Paul Röttger. 2023.
\newblock \href {https://doi.org/10.48550/arXiv.2303.04222} {{SemEval}-2023
  {Task} 10: {Explainable} {Detection} of {Online} {Sexism}}.
\newblock In \emph{Proceedings of the 17th {{International Workshop}} on
  {{Semantic Evaluation}} ({{SemEval-2023}})}. {Association for Computational
  Linguistics}.

\bibitem[{Kov{\'a}cs et~al.(2021)Kov{\'a}cs, Alonso, and
  Saini}]{kovacs2021challenges}
Gy{\"o}rgy Kov{\'a}cs, Pedro Alonso, and Rajkumar Saini. 2021.
\newblock Challenges of hate speech detection in social media: Data scarcity,
  and leveraging external resources.
\newblock \emph{SN Computer Science}, 2(2):95.

\bibitem[{Kurrek et~al.(2020)Kurrek, Saleem, and
  Ruths}]{kurrek-etal-2020-towards}
Jana Kurrek, Haji~Mohammad Saleem, and Derek Ruths. 2020.
\newblock \href {https://doi.org/10.18653/v1/2020.alw-1.17} {Towards a
  comprehensive taxonomy and large-scale annotated corpus for online slur
  usage}.
\newblock In \emph{Proceedings of the Fourth Workshop on Online Abuse and
  Harms}, pages 138--149, Online. Association for Computational Linguistics.

\bibitem[{Lewandowska-Tomaszczyk et~al.(2023)Lewandowska-Tomaszczyk,
  {\v{Z}}itnik, Liebeskind, Valunaite~Oleskevicien{\.e}, B{\k{a}}czkowska,
  Wilson, Trojszczak, Bra{\v{c}}, Filipi{\'c}, Ostro{\v{s}}ki~Ani{\'c}
  et~al.}]{lewandowska2023annotation}
Barbara Lewandowska-Tomaszczyk, Slavko {\v{Z}}itnik, Chaya Liebeskind, Giedre
  Valunaite~Oleskevicien{\.e}, Anna B{\k{a}}czkowska, Paul~A Wilson, Marcin
  Trojszczak, Ivana Bra{\v{c}}, Lobel Filipi{\'c}, Ana Ostro{\v{s}}ki~Ani{\'c},
  and 1 others. 2023.
\newblock Annotation scheme and evaluation: The case of offensive language.
\newblock \emph{Rasprave: {\v{C}}asopis Instituta za hrvatski jezik i
  jezikoslovlje}, 49(1).

\bibitem[{Li et~al.(2024)Li, Fan, Atreja, and Hemphill}]{li2024hot}
Lingyao Li, Lizhou Fan, Shubham Atreja, and Libby Hemphill. 2024.
\newblock \href {https://dl.acm.org/doi/10.1145/3643829} {“hot” chatgpt:
  The promise of chatgpt in detecting and discriminating hateful, offensive,
  and toxic comments on social media}.
\newblock \emph{ACM Transactions on the Web}, 18(2):1--36.

\bibitem[{Liu et~al.(2019)Liu, Ott, Goyal, Du, Joshi, Chen, Levy, Lewis,
  Zettlemoyer, and Stoyanov}]{DBLP:journals/corr/abs-1907-11692}
Yinhan Liu, Myle Ott, Naman Goyal, Jingfei Du, Mandar Joshi, Danqi Chen, Omer
  Levy, Mike Lewis, Luke Zettlemoyer, and Veselin Stoyanov. 2019.
\newblock \href {https://arxiv.org/abs/1907.11692} {Roberta: {A} robustly
  optimized {BERT} pretraining approach}.
\newblock \emph{CoRR}, abs/1907.11692.

\bibitem[{Logacheva et~al.(2022)Logacheva, Dementieva, Ustyantsev, Moskovskiy,
  Dale, Krotova, Semenov, and Panchenko}]{logacheva-etal-2022-paradetox}
Varvara Logacheva, Daryna Dementieva, Sergey Ustyantsev, Daniil Moskovskiy,
  David Dale, Irina Krotova, Nikita Semenov, and Alexander Panchenko. 2022.
\newblock \href {https://doi.org/10.18653/v1/2022.acl-long.469} {{P}ara{D}etox:
  Detoxification with parallel data}.
\newblock In \emph{Proceedings of the 60th Annual Meeting of the Association
  for Computational Linguistics (Volume 1: Long Papers)}, pages 6804--6818,
  Dublin, Ireland. Association for Computational Linguistics.

\bibitem[{Mandl et~al.(2019)Mandl, Modha, Majumder, Patel, Dave, Mandlia, and
  Patel}]{mandl2019overview}
Thomas Mandl, Sandip Modha, Prasenjit Majumder, Daksh Patel, Mohana Dave,
  Chintak Mandlia, and Aditya Patel. 2019.
\newblock Overview of the hasoc track at fire 2019: Hate speech and offensive
  content identification in indo-european languages.
\newblock In \emph{Proceedings of the 11th annual meeting of the Forum for
  Information Retrieval Evaluation}, pages 14--17.

\bibitem[{Mathew et~al.(2021)Mathew, Saha, Yimam, Biemann, Goyal, and
  Mukherjee}]{mathew2021hatexplain}
Binny Mathew, Punyajoy Saha, Seid~Muhie Yimam, Chris Biemann, Pawan Goyal, and
  Animesh Mukherjee. 2021.
\newblock Hatexplain: A benchmark dataset for explainable hate speech
  detection.
\newblock In \emph{Proceedings of the AAAI Conference on Artificial
  Intelligence}, volume~35, pages 14867--14875.

\bibitem[{Mazari et~al.(2024)Mazari, Boudoukhani, and Djeffal}]{mazari2024bert}
Ahmed~Cherif Mazari, Nesrine Boudoukhani, and Abdelhamid Djeffal. 2024.
\newblock Bert-based ensemble learning for multi-aspect hate speech detection.
\newblock \emph{Cluster Computing}, 27(1):325--339.

\bibitem[{Mishra and Mishra(2019)}]{mishra20193idiots}
Shubhanshu Mishra and Sudhanshu Mishra. 2019.
\newblock 3idiots at hasoc 2019: Fine-tuning transformer neural networks for
  hate speech identification in indo-european languages.
\newblock In \emph{FIRE (working notes)}, pages 208--213.

\bibitem[{Mollas et~al.(2022)Mollas, Chrysopoulou, Karlos, and
  Tsoumakas}]{mollas2022ethos}
Ioannis Mollas, Zoe Chrysopoulou, Stamatis Karlos, and Grigorios Tsoumakas.
  2022.
\newblock Ethos: a multi-label hate speech detection dataset.
\newblock \emph{Complex \& Intelligent Systems}, 8(6):4663--4678.

\bibitem[{Muti and
  Barr{\'o}n-Cede{\~n}o(2022)}]{muti-barron-cedeno-2022-checkpoint}
Arianna Muti and Alberto Barr{\'o}n-Cede{\~n}o. 2022.
\newblock \href {https://doi.org/10.18653/v1/2022.acl-srw.37} {A checkpoint on
  multilingual misogyny identification}.
\newblock In \emph{Proceedings of the 60th Annual Meeting of the Association
  for Computational Linguistics: Student Research Workshop}, pages 454--460,
  Dublin, Ireland. Association for Computational Linguistics.

\bibitem[{Nasir et~al.(2025)Nasir, Sharma, Jaidka, and
  Ahmed}]{DBLP:conf/iccsa/NasirSJA25}
Ahmad Nasir, Aadish Sharma, Kokil Jaidka, and Saifuddin Ahmed. 2025.
\newblock \href {https://doi.org/10.1007/978-3-031-96962-1\_2} {Llms and
  finetuning: Benchmarking cross-domain performance for hate speech detection}.
\newblock In \emph{Computational Science and Its Applications - {ICCSA} 2025 -
  25th International Conference, Istanbul, Turkey, June 30 - July 3, 2025,
  Proceedings, Part {III}}, volume 15650 of \emph{Lecture Notes in Computer
  Science}, pages 17--34. Springer.

\bibitem[{Naveed et~al.(2025)Naveed, Khan, Qiu, Saqib, Anwar, Usman, Akhtar,
  Barnes, and Mian}]{DBLP:journals/tist/NaveedKQSAUABM25}
Humza Naveed, Asad~Ullah Khan, Shi Qiu, Muhammad Saqib, Saeed Anwar, Muhammad
  Usman, Naveed Akhtar, Nick Barnes, and Ajmal Mian. 2025.
\newblock \href {https://doi.org/10.1145/3744746} {A comprehensive overview of
  large language models}.
\newblock \emph{{ACM} Trans. Intell. Syst. Technol.}, 16(5):106:1--106:72.

\bibitem[{Ousidhoum et~al.(2019)Ousidhoum, Lin, Zhang, Song, and
  Yeung}]{ousidhoum2019multilingual}
Nedjma Ousidhoum, Zizheng Lin, Hongming Zhang, Yangqiu Song, and Dit-Yan Yeung.
  2019.
\newblock Multilingual and multi-aspect hate speech analysis.
\newblock \emph{arXiv preprint arXiv:1908.11049}.

\bibitem[{Pamungkas et~al.(2020{\natexlab{a}})Pamungkas, Basile, and
  Patti}]{pamungkas2020you}
Endang~Wahyu Pamungkas, Valerio Basile, and Viviana Patti. 2020{\natexlab{a}}.
\newblock Do you really want to hurt me? predicting abusive swearing in social
  media.
\newblock In \emph{Proceedings of the Twelfth Language Resources and Evaluation
  Conference}, pages 6237--6246.

\bibitem[{Pamungkas et~al.(2020{\natexlab{b}})Pamungkas, Basile, and
  Patti}]{pamungkas2020misogyny}
Endang~Wahyu Pamungkas, Valerio Basile, and Viviana Patti. 2020{\natexlab{b}}.
\newblock Misogyny detection in twitter: a multilingual and cross-domain study.
\newblock \emph{Information processing \& management}, 57(6):102360.

\bibitem[{Pan et~al.(2024)Pan, Garc{\'\i}a-D{\'\i}az, and
  Valencia-Garc{\'\i}a}]{pan2024comparing}
Ronghao Pan, Jos{\'e}~Antonio Garc{\'\i}a-D{\'\i}az, and Rafael
  Valencia-Garc{\'\i}a. 2024.
\newblock Comparing fine-tuning, zero and few-shot strategies with large
  language models in hate speech detection in english.
\newblock \emph{CMES-Computer Modeling in Engineering \& Sciences}, 140(3).

\bibitem[{Pavlopoulos et~al.(2022)Pavlopoulos, Laugier, Xenos, Sorensen, and
  Androutsopoulos}]{pavlopoulos-etal-2022-acl}
John Pavlopoulos, L{\'e}o Laugier, Alexandros Xenos, Jeffrey Sorensen, and Ion
  Androutsopoulos. 2022.
\newblock From the detection of toxic spans in online discussions to the
  analysis of toxic-to-civil transfer.
\newblock In \emph{Proceedings of the 60th Annual Meeting of the Association
  for Computational Linguistics (ACL 2022).}, Dublin, Ireland. Association for
  Computational Linguistics.

\bibitem[{Pavlopoulos et~al.(2020)Pavlopoulos, Sorensen, Dixon, Thain, and
  Androutsopoulos}]{pavlopoulos2020toxicity}
John Pavlopoulos, Jeffrey Sorensen, Lucas Dixon, Nithum Thain, and Ion
  Androutsopoulos. 2020.
\newblock Toxicity detection: Does context really matter?
\newblock \emph{arXiv preprint arXiv:2006.00998}.

\bibitem[{Pavlopoulos et~al.(2021)Pavlopoulos, Sorensen, Laugier, and
  Androutsopoulos}]{pavlopoulos-etal-2021-semeval}
John Pavlopoulos, Jeffrey Sorensen, L{\'e}o Laugier, and Ion Androutsopoulos.
  2021.
\newblock \href {https://doi.org/10.18653/v1/2021.semeval-1.6}
  {{S}em{E}val-2021 task 5: Toxic spans detection}.
\newblock In \emph{Proceedings of the 15th International Workshop on Semantic
  Evaluation (SemEval-2021)}, pages 59--69, Online. Association for
  Computational Linguistics.

\bibitem[{Piot et~al.(2024)Piot, Mart{\'\i}n-Rodilla, and
  Parapar}]{piot2024metahate}
Paloma Piot, Patricia Mart{\'\i}n-Rodilla, and Javier Parapar. 2024.
\newblock Metahate: A dataset for unifying efforts on hate speech detection.
\newblock In \emph{Proceedings of the International AAAI Conference on Web and
  Social Media}, volume~18, pages 2025--2039.

\bibitem[{Poletto et~al.(2021)Poletto, Basile, Sanguinetti, Bosco, and
  Patti}]{DBLP:journals/lre/PolettoBSBP21}
Fabio Poletto, Valerio Basile, Manuela Sanguinetti, Cristina Bosco, and Viviana
  Patti. 2021.
\newblock \href {https://doi.org/10.1007/S10579-020-09502-8} {Resources and
  benchmark corpora for hate speech detection: a systematic review}.
\newblock \emph{Lang. Resour. Evaluation}, 55(2):477--523.

\bibitem[{Qian et~al.(2019)Qian, Bethke, Liu, Belding, and
  Wang}]{qian2019benchmark}
Jing Qian, Anna Bethke, Yinyin Liu, Elizabeth Belding, and William~Yang Wang.
  2019.
\newblock A benchmark dataset for learning to intervene in online hate speech.
\newblock \emph{arXiv preprint arXiv:1909.04251}.

\bibitem[{{Qwen Team}(2025)}]{qwen3technicalreport}
{Qwen Team}. 2025.
\newblock \href {https://arxiv.org/abs/2505.09388} {Qwen3 technical report}.
\newblock \emph{Preprint}, arXiv:2505.09388.

\bibitem[{Ramponi et~al.(2022)Ramponi, Testa, Tonelli, and
  Jezek}]{ramponi2022addressing}
Alan Ramponi, Benedetta Testa, Sara Tonelli, and Elisabetta Jezek. 2022.
\newblock Addressing religious hate online: from taxonomy creation to automated
  detection.
\newblock \emph{PeerJ Computer Science}, 8:e1128.

\bibitem[{Rizwan et~al.(2025)Rizwan, Yimam, Dementieva, Skupin, Fischer,
  Moskovskiy, Borkar, Geislinger, Saha, Roy, Semmann, Panchenko, Biemann, and
  Mukherjee}]{rizwan-etal-2025-hateprism}
Naquee Rizwan, Seid~Muhie Yimam, Daryna Dementieva, Dr.~Florian Skupin, Tim
  Fischer, Daniil Moskovskiy, Aarushi~Ajay Borkar, Robert Geislinger, Punyajoy
  Saha, Sarthak Roy, Martin Semmann, Alexander Panchenko, Chris Biemann, and
  Animesh Mukherjee. 2025.
\newblock \href {https://doi.org/10.18653/v1/2025.findings-acl.824}
  {{H}ate{PRISM}: Policies, platforms, and research integration. advancing
  {NLP} for hate speech proactive mitigation}.
\newblock In \emph{Findings of the Association for Computational Linguistics:
  ACL 2025}, pages 16008--16022, Vienna, Austria. Association for Computational
  Linguistics.

\bibitem[{R{\"o}ttger et~al.(2021)R{\"o}ttger, Vidgen, Nguyen, Waseem,
  Margetts, and Pierrehumbert}]{rottger-etal-2021-hatecheck}
Paul R{\"o}ttger, Bertie Vidgen, Dong Nguyen, Zeerak Waseem, Helen Margetts,
  and Janet Pierrehumbert. 2021.
\newblock \href {https://doi.org/10.18653/v1/2021.acl-long.4} {{H}ate{C}heck:
  Functional tests for hate speech detection models}.
\newblock In \emph{Proceedings of the 59th Annual Meeting of the Association
  for Computational Linguistics and the 11th International Joint Conference on
  Natural Language Processing (Volume 1: Long Papers)}, pages 41--58, Online.
  Association for Computational Linguistics.

\bibitem[{Roy et~al.(2023)Roy, Harshvardhan, Mukherjee, and
  Saha}]{roy2023probing}
Sarthak Roy, Ashish Harshvardhan, Animesh Mukherjee, and Punyajoy Saha. 2023.
\newblock Probing llms for hate speech detection: strengths and
  vulnerabilities.
\newblock In \emph{Findings of the association for computational linguistics:
  EMNLP 2023}, pages 6116--6128.

\bibitem[{Sachdeva et~al.(2022)Sachdeva, Barreto, Bacon, Sahn, von Vacano, and
  Kennedy}]{sachdeva-etal-2022-measuring}
Pratik~S. Sachdeva, Renata Barreto, Geoff Bacon, Alexander Sahn, Claudia von
  Vacano, and Chris Kennedy. 2022.
\newblock \href {https://aclanthology.org/2022.nlperspectives-1.11} {The
  measuring hate speech corpus: Leveraging rasch measurement theory for data
  perspectivism}.
\newblock In \emph{Proceedings of the 1st Workshop on Perspectivist Approaches
  to NLP @LREC2022}, pages 83--94, Marseille, France. European Language
  Resources Association.

\bibitem[{Sai and Sharma(2020)}]{DBLP:conf/fire/SaiS20}
Siva Sai and Yashvardhan Sharma. 2020.
\newblock \href {https://ceur-ws.org/Vol-2826/T2-32.pdf}
  {Siva@hasoc-dravidian-codemix-fire-2020: Multilingual offensive speech
  detection in code-mixed and romanized text}.
\newblock In \emph{Working Notes of {FIRE} 2020 - Forum for Information
  Retrieval Evaluation, Hyderabad, India, December 16-20, 2020}, volume 2826 of
  \emph{{CEUR} Workshop Proceedings}, pages 336--343. CEUR-WS.org.

\bibitem[{Salminen et~al.(2018)Salminen, Almerekhi, Milenkovi{\'c}, Jung, An,
  Kwak, and Jansen}]{salminen2018anatomy}
Joni Salminen, Hind Almerekhi, Milica Milenkovi{\'c}, Soon-gyo Jung, Jisun An,
  Haewoon Kwak, and Bernard Jansen. 2018.
\newblock Anatomy of online hate: developing a taxonomy and machine learning
  models for identifying and classifying hate in online news media.
\newblock In \emph{Proceedings of the International AAAI Conference on Web and
  Social Media}, volume~12.

\bibitem[{Sarker et~al.(2023)Sarker, Turzo, Dong, and
  Bosu}]{sarker2023automated}
Jaydeb Sarker, Asif~Kamal Turzo, Ming Dong, and Amiangshu Bosu. 2023.
\newblock Automated identification of toxic code reviews using toxicr.
\newblock \emph{ACM Transactions on Software Engineering and Methodology},
  32(5):1--32.

\bibitem[{Sen et~al.(2024)Sen, Das, and
  Sen}]{DBLP:journals/corr/abs-2405-01577}
Tanmay Sen, Ansuman Das, and Mrinmay Sen. 2024.
\newblock \href {https://doi.org/10.48550/ARXIV.2405.01577} {Hatetinyllm : Hate
  speech detection using tiny large language models}.
\newblock \emph{CoRR}, abs/2405.01577.

\bibitem[{Sorensen et~al.(2023)Sorensen, Korre, Pavlopoulos, Tomanek, Thain,
  Dixon, and Laugier}]{sorensen-etal-2023-juage}
Jeffrey Sorensen, Katerina Korre, John Pavlopoulos, Katrin Tomanek, Nithum
  Thain, Lucas Dixon, and L{\'e}o Laugier. 2023.
\newblock \href {https://doi.org/10.18653/v1/2023.semeval-1.166} {{JUAGE} at
  {S}em{E}val-2023 task 10: Parameter efficient classification}.
\newblock In \emph{Proceedings of the 17th International Workshop on Semantic
  Evaluation (SemEval-2023)}, pages 1195--1203, Toronto, Canada. Association
  for Computational Linguistics.

\bibitem[{Toraman et~al.(2022)Toraman, {\c{S}}ahinu{\c{c}}, and
  Yilmaz}]{toraman-etal-2022-large}
Cagri Toraman, Furkan {\c{S}}ahinu{\c{c}}, and Eyup Yilmaz. 2022.
\newblock \href {https://aclanthology.org/2022.lrec-1.238} {Large-scale hate
  speech detection with cross-domain transfer}.
\newblock In \emph{Proceedings of the Thirteenth Language Resources and
  Evaluation Conference}, pages 2215--2225, Marseille, France. European
  Language Resources Association.

\bibitem[{Vidgen and Derczynski(2020)}]{vidgen2020directions}
Bertie Vidgen and Leon Derczynski. 2020.
\newblock Directions in abusive language training data, a systematic review:
  Garbage in, garbage out.
\newblock \emph{Plos one}, 15(12):e0243300.

\bibitem[{Vidgen et~al.(2020)Vidgen, Hale, Guest, Margetts, Broniatowski,
  Talat, Botelho, Hall, and Tromble}]{vidgen2020detecting}
Bertie Vidgen, Scott~A Hale, Ella Guest, Helen Margetts, David Broniatowski,
  Zeerak Talat, Austin Botelho, Matthew Hall, and Rebekah Tromble. 2020.
\newblock Detecting east asian prejudice on social media.
\newblock In \emph{Proceedings of the fourth workshop on online abuse and
  harms}, pages 162--172.

\bibitem[{Vidgen et~al.(2021)Vidgen, Nguyen, Margetts, Rossini, and
  Tromble}]{vidgen2021introducing}
Bertie Vidgen, Dong Nguyen, Helen Margetts, Patricia Rossini, and Rebekah
  Tromble. 2021.
\newblock Introducing cad: the contextual abuse dataset.

\bibitem[{Wang et~al.(2020)Wang, Liu, Ouyang, and Sun}]{wang-etal-2020-galileo}
Shuohuan Wang, Jiaxiang Liu, Xuan Ouyang, and Yu~Sun. 2020.
\newblock \href {https://doi.org/10.18653/v1/2020.semeval-1.189} {Galileo at
  {S}em{E}val-2020 task 12: Multi-lingual learning for offensive language
  identification using pre-trained language models}.
\newblock In \emph{Proceedings of the Fourteenth Workshop on Semantic
  Evaluation}, pages 1448--1455, Barcelona (online). International Committee
  for Computational Linguistics.

\bibitem[{Waseem(2016)}]{waseem:2016:NLPandCSS}
Zeerak Waseem. 2016.
\newblock \href {http://aclweb.org/anthology/W16-5618} {Are you a racist or am
  i seeing things? annotator influence on hate speech detection on twitter}.
\newblock In \emph{Proceedings of the First Workshop on NLP and Computational
  Social Science}, pages 138--142, Austin, Texas. Association for Computational
  Linguistics.

\bibitem[{Waseem and Hovy(2016)}]{waseem-hovy:2016:N16-2}
Zeerak Waseem and Dirk Hovy. 2016.
\newblock \href {http://www.aclweb.org/anthology/N16-2013} {Hateful symbols or
  hateful people? predictive features for hate speech detection on twitter}.
\newblock In \emph{Proceedings of the NAACL Student Research Workshop}, pages
  88--93, San Diego, California. Association for Computational Linguistics.

\bibitem[{Wiedemann et~al.(2020)Wiedemann, Yimam, and
  Biemann}]{wiedemann-etal-2020-uhh}
Gregor Wiedemann, Seid~Muhie Yimam, and Chris Biemann. 2020.
\newblock \href {https://doi.org/10.18653/v1/2020.semeval-1.213} {{UHH}-{LT} at
  {S}em{E}val-2020 task 12: Fine-tuning of pre-trained transformer networks for
  offensive language detection}.
\newblock In \emph{Proceedings of the Fourteenth Workshop on Semantic
  Evaluation}, pages 1638--1644, Barcelona (online). International Committee
  for Computational Linguistics.

\bibitem[{Yu et~al.(2022)Yu, Blanco, and Hong}]{yu-etal-2022-hate}
Xinchen Yu, Eduardo Blanco, and Lingzi Hong. 2022.
\newblock \href {https://doi.org/10.18653/v1/2022.naacl-main.433} {Hate speech
  and counter speech detection: Conversational context does matter}.
\newblock In \emph{Proceedings of the 2022 Conference of the North American
  Chapter of the Association for Computational Linguistics: Human Language
  Technologies}, pages 5918--5930, Seattle, United States. Association for
  Computational Linguistics.

\bibitem[{Zampieri et~al.(2019)Zampieri, Malmasi, Nakov, Rosenthal, Farra, and
  Kumar}]{zampieri2019predicting}
Marcos Zampieri, Shervin Malmasi, Preslav Nakov, Sara Rosenthal, Noura Farra,
  and Ritesh Kumar. 2019.
\newblock Predicting the type and target of offensive posts in social media.
\newblock \emph{arXiv preprint arXiv:1902.09666}.

\bibitem[{Zhang et~al.(2024)Zhang, Deng, Liu, Pan, and
  Bing}]{zhang-etal-2024-sentiment}
Wenxuan Zhang, Yue Deng, Bing Liu, Sinno Pan, and Lidong Bing. 2024.
\newblock \href {https://doi.org/10.18653/v1/2024.findings-naacl.246}
  {Sentiment analysis in the era of large language models: A reality check}.
\newblock In \emph{Findings of the Association for Computational Linguistics:
  NAACL 2024}, pages 3881--3906, Mexico City, Mexico. Association for
  Computational Linguistics.

\bibitem[{Zhou(2023)}]{zhou-2023-pinganlifeinsurance}
Mengyuan Zhou. 2023.
\newblock \href {https://doi.org/10.18653/v1/2023.semeval-1.304}
  {{P}ing{A}n{L}ife{I}nsurance at {S}em{E}val-2023 task 10: Using multi-task
  learning to better detect online sexism}.
\newblock In \emph{Proceedings of the 17th International Workshop on Semantic
  Evaluation (SemEval-2023)}, pages 2188--2192, Toronto, Canada. Association
  for Computational Linguistics.

\bibitem[{Zhu et~al.(2021)Zhu, Lin, Zhang, Sun, Li, Lin, Dang, and
  Xu}]{zhu-etal-2021-hitsz}
Qinglin Zhu, Zijie Lin, Yice Zhang, Jingyi Sun, Xiang Li, Qihui Lin, Yixue
  Dang, and Ruifeng Xu. 2021.
\newblock \href {https://doi.org/10.18653/v1/2021.semeval-1.63} {{HITSZ}-{HLT}
  at {S}em{E}val-2021 task 5: Ensemble sequence labeling and span boundary
  detection for toxic span detection}.
\newblock In \emph{Proceedings of the 15th International Workshop on Semantic
  Evaluation (SemEval-2021)}, pages 521--526, Online. Association for
  Computational Linguistics.

\bibitem[{Zhu et~al.(2025)Zhu, Zhang, Haq, Hui, and Tyson}]{zhu2023can}
Yiming Zhu, Peixian Zhang, Ehsan-Ul Haq, Pan Hui, and Gareth Tyson. 2025.
\newblock \href {https://link.springer.com/chapter/10.1007/978-3-031-78548-1_2}
  {Exploring the capability of chatgpt to reproduce human labels for social
  computing tasks}.
\newblock In \emph{Social Networks Analysis and Mining}, pages 13--22. Springer
  Nature Switzerland.

\end{thebibliography}

\appendix

\section{All Prompts}
\label{app:prompts}
Here we list all prompts used in training and testing. They are categorized first by task type, followed by specific prompts for individual datasets. We omit the generic message before the input text as well as newlines for brevity. All of the prompts are included in our GitHub repository. 

\subsection{Generic Prompts}

\paragraph{Binary Classification}
\begin{quote}
\textbf{User:} Should the text be classified as \{\texttt{category}\}? Answer with `yes' or `no'.
\end{quote}
\paragraph{Multiclass Classification}
\begin{quote}
\textbf{User:} Which category does the text belong to? Categories: \{\texttt{categories}\}    
\end{quote}
The \texttt{categories} are a newline-separated list, with `A. ', `B. ', `C. ', etc. added before each entry. For tasks where the category is a specific target, we use this prompt instead: 
\begin{quote}    
\textbf{User:} Which of the following groups does the text target? Groups: \\ \{\texttt{categories}\}  
\end{quote}
\paragraph{Multilabel Classification}
\begin{quote}
\textbf{User:} Which of the following categories does the text belong to? You can choose multiple, but keep the order of the categories. Categories: \{\texttt{categories}\}    
\end{quote}
The \texttt{categories} are a newline-separated list, with `A. ', `B. ', `C. ', etc. added before each entry. For tasks where the category is a specific target (and can be multiple targets), we use this prompt instead: 
\begin{quote}    
\textbf{User:} Which of the following groups does the text target? You can choose multiple, but keep the order of the groups. Groups: \{\texttt{categories}\}  
\end{quote}
\subsection{Specific Prompts}
\paragraph{Counter Speech}
For CONAN and Multitarget-CONAN, we use:
\begin{quote}
    \textbf{User:} Provide a counter-narrative for the text.
\end{quote}
For DialoCONAN, we use:
\begin{quote}
    \textbf{User:} Have a dialogue with the author of the original text. Provide and maintain a counter-narrative throughout the dialogue.    
\end{quote}

\paragraph{Ethos}
Ethos has two annotations which required prompts that did not fit into our generic prompting strategies.
\begin{quote}
    \textbf{User:} Does the comment incite violence? Answer `yes' or `no'.
\end{quote}

\begin{quote}
    \textbf{User:} Is the text targeting a specific individual (directed) or a group/class of people (generalized)? Answer `directed' or `generalized'.
\end{quote}

\paragraph{Gab Hate Corpus}
One of the tasks in the GHC is to annotate whether the hateful content is said in an implicit or explicit way. 
\begin{quote}
    \textbf{User:} Is the rhetoric in the text explicit or implicit? Answer with `explicit' or `implicit'.
\end{quote}

\paragraph{HateExplain} 
Although we do not use it for testing, we include the task of identifying parts of the text that are hateful or offensive. This prompt is only given if the text was classified as such.
\begin{quote}
\textbf{User:} Identify all of the words in the text that could be used to justify the given classification. Write them in a space-separated list.
\end{quote}

\paragraph{Implicit Hate}
For implicit hate, we only use one of the tasks, which is to infer the speaker's meaning from their comment.
\begin{quote}
    \textbf{User:} What is the implied statement of this post?
\end{quote}

\paragraph{Intervene}
The intervention dataset includes examples on multiple lines, with each line labeled with an index. The task is to first identify which lines constitute hate speech, then to form an intervention. Each example had 3 possible interventions. The detection prompt is as follows:
\begin{quote}
\textbf{User:} Which posts or comments in this conversation are hate speech? Refer to their numbers, e.g. 1, 2, 3, or n/a if none of the posts or comments are hate speech.
\end{quote}
The intervention prompts are as follows:
\begin{quote}
\textbf{User:} How would you respond to intervene? Limit your response to 150 characters.
\end{quote}
\begin{quote}
\textbf{User:} Come up with another response to intervene.
\end{quote}
\begin{quote}
\textbf{User:} Come up with a third response to intervene.
\end{quote}

\paragraph{ParaDetox}
\begin{quote}
\textbf{User:}  Rephrase the text to make it less toxic and more neutral.
\end{quote}

\paragraph{SWAD}
\begin{quote}
    \textbf{User:} Is the word marked in bold (<b></b>) used in an abusive manner?
\end{quote}

\paragraph{ToxicSpans}
\begin{quote}
    \textbf{User:} Identify the part(s) of the given text that is/are considered toxic. If there are multiple spans, separate them with semicolons. Write N/A if the text is not toxic.
\end{quote}

\section{Hyperparameters} \label{app:hyperparams}
In Table \ref{tab:hyperparams}, we list the hyperparameters specified in training. 
\begin{table}[]
    \centering
    \begin{tabular}{lr} \toprule
    Parameter & Value \\ \midrule
    Batch size & 64 \\
    Learning rate & 1e-4 \\
    Weight decay & 1e-3 \\
    Warmup ratio & 1e-2 \\
    Epochs & 3 \\
    LoRA rank & 32 \\
    LoRA alpha & 32 \\
    LoRA dropout & 0 \\
    LoRA bias & None \\
    Optimizer & AdamW 8-bit \\
    Scheduler & Cosine \\
    Quantization & 4-bit NF4 double \\
    
    \bottomrule
    \end{tabular}
    \caption{Hyperparameters used in training.}
    \label{tab:hyperparams}
\end{table}

\section{Datasets Labels Statistics}
\label{app:labels_statistics}

\begin{figure}[!htp]
    \centering
    \includegraphics[width=\linewidth]{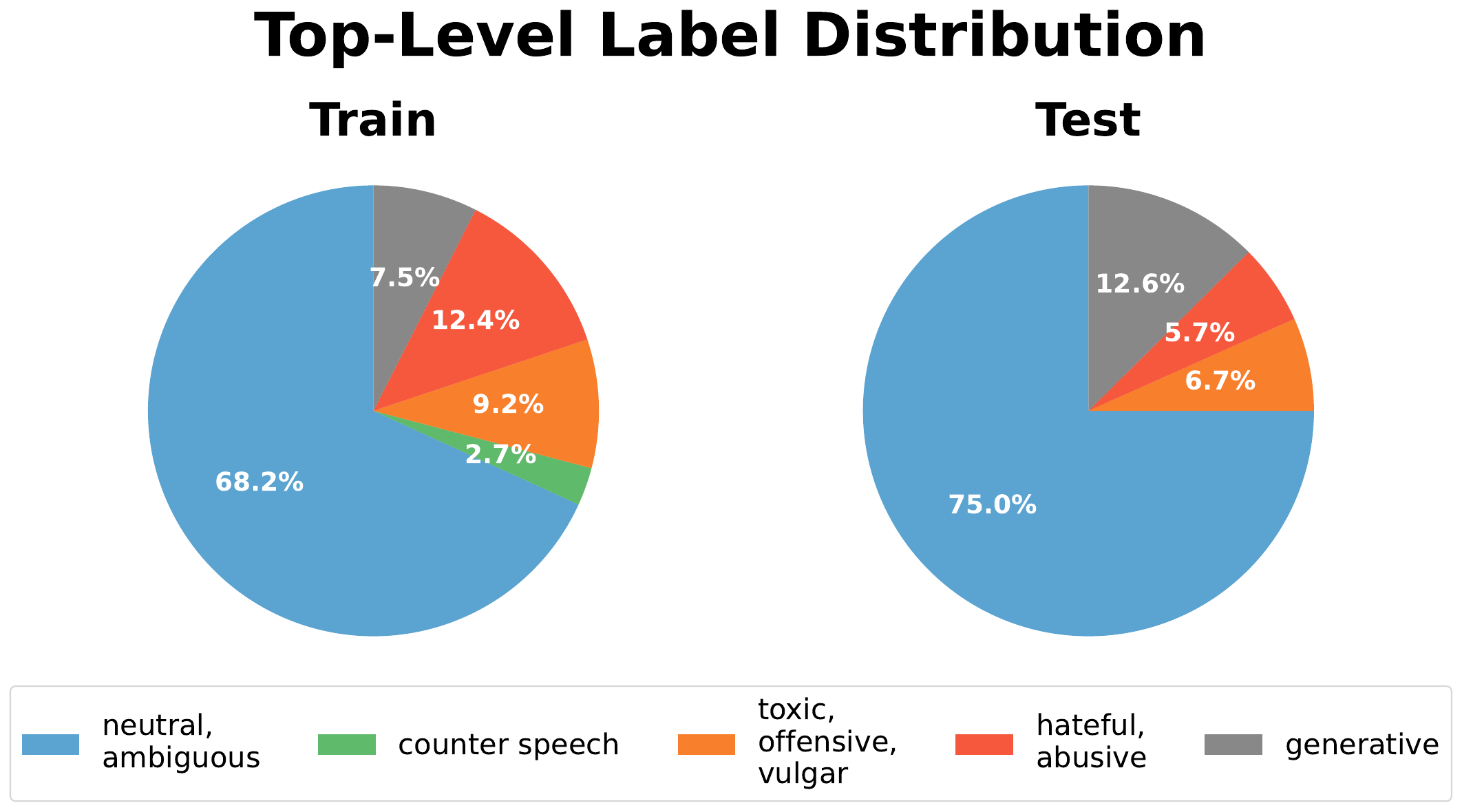}
    \caption{Top-level label distribution of all datasets.}
    \label{fig:top_labels}
\end{figure}

Here, we give a general overview of the label distribution of the data.
Figure \ref{fig:top_labels} shows the top-level labels of all datasets, grouped into 5 categories. The `generative' category consists of the datasets with generative tasks: ToxicSpans, ParaDetox, ImplicitHate, Intervene, and the CONAN datasets. If these were discriminative tasks, the examples would largely fall under the toxic or hateful categories, as the tasks presume the text is toxic or hateful. 

\begin{figure}[!htp]
    \centering
    \includegraphics[width=\linewidth]{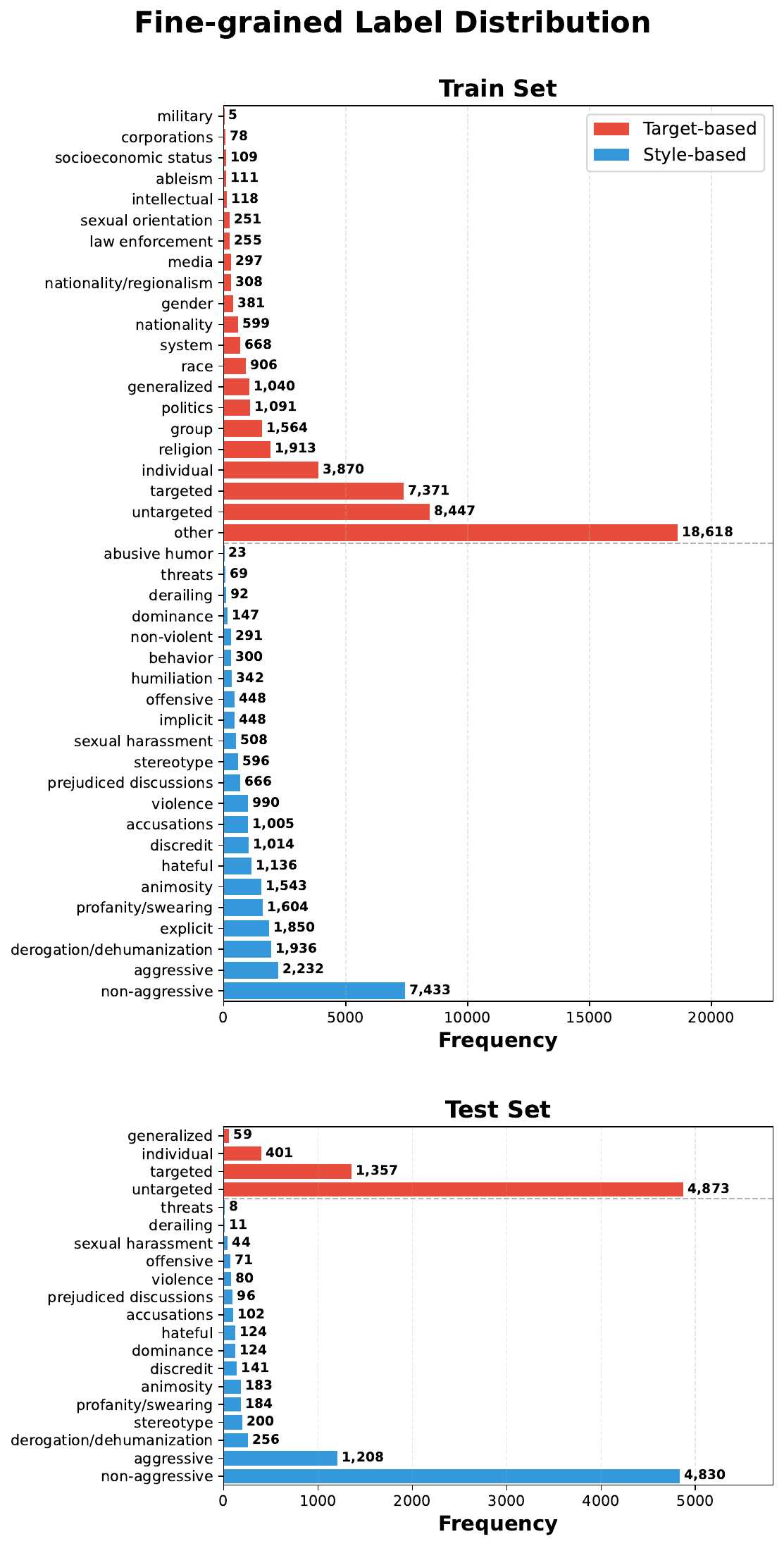}
    \caption{Fine-grained label distribution of the hierarchical datasets.}
    \label{fig:nested_labels}
\end{figure}

In Figure \ref{fig:nested_labels}, we break down the datasets with finer-grained labels, not included in Figure \ref{fig:top_labels}. We find that the finer-grained labels can be broken into two types: target-based and style-based labels. Target-based labels refer to labeling the target of the hateful or offensive speech. Style-based labels refer to the manner in which it is hateful or offensive. These labels are not mutually exclusive, nor are the labels within each category.
We group similar labels (e.g., sexism, misogyny, and gender) for brevity; however, there may be overlap between some label groups. Overall, we see that target-based labels are more frequent in the training set, but the types are more evenly distributed in the test set. We also see that generic labels, such as whether an example is targeted or not, are more frequent.
The large `other' group mostly comes from the Gab Hate Corpus, in which examples were annotated for a variety of targets, so other could either refer to untargeted hate or targeted hate that does not fall into the included categories.

\section{Licensing of Resources}
\label{sec:app_licenses}

Below is an overview of the licenses associated with each resource used in this work (Table~\ref{tab:overview-license}).

\begin{table*}[ht!]
\centering
\scriptsize
\begin{tabular}{p{2.2cm}p{2cm}p{10.2cm}}
\toprule
Resource & License & Homepage  \\ 
\midrule
CAD & CC BY 4.0 & \href{https://github.com/dongpng/cad_naacl2021}{https://github.com/dongpng/cad\_naacl2021} \\
CAT-LARGE & Apache-2.0 & \href{https://github.com/ipavlopoulos/context_toxicity}{https://github.com/ipavlopoulos/context\_toxicity} \\
CONAN & Research Only & \href{https://github.com/marcoguerini/CONAN}{https://github.com/marcoguerini/CONAN}\\
ConvAbuse & CC BY 4.0 & \href{https://github.com/amandacurry/convabuse}{https://github.com/amandacurry/convabuse}\\
DialoCONAN & Research Only & \href{https://github.com/marcoguerini/CONAN}{https://github.com/marcoguerini/CONAN} \\
EAP & CC BY 4.0 & \href{https://zenodo.org/records/3816667}{https://zenodo.org/records/3816667}\\
ETHOS & GNU GPLv3 & \href{https://github.com/intelligence-csd-auth-gr/Ethos-Hate-Speech-Dataset}{https://github.com/intelligence-csd-auth-gr/Ethos-Hate-Speech-Dataset}\\
FoxCom & MIT & \href{https://github.com/sjtuprog/fox-news-comments}{https://github.com/sjtuprog/fox-news-comments}\\
GHC & CC BY 4.0 & \href{https://osf.io/edua3/overview}{https://osf.io/edua3/overview}\\
HSOL & MIT & \href{https://huggingface.co/datasets/tdavidson/hate_speech_offensive}{https://huggingface.co/datasets/tdavidson/hate\_speech\_offensive}\\
ImplicitHate & MIT & \href{https://github.com/SALT-NLP/implicit-hate}{https://github.com/SALT-NLP/implicit-hate} \\
Intervene & CC BY-NC 4.0 & \href{https://github.com/jing-qian/A-Benchmark-Dataset-for-Learning-to-Intervene-in-Online-Hate-Speech}{https://github.com/jing-qian/A-Benchmark-Dataset-for-Learning-to-Intervene-in-Online-Hate-Speech}\\
LargeScaleAbuse & CC BY & \href{https://github.com/ENCASEH2020/crowdflower-platform}{https://github.com/ENCASEH2020/crowdflower-platform}\\
LargeScaleXDomain & CC BY-NC 4.0 & \href{https://github.com/avaapm/hatespeech}{https://github.com/avaapm/hatespeech}\\
MeasuringHate & CC BY 4.0 & \href{https://huggingface.co/datasets/ucberkeley-dlab/measuring-hate-speech}{https://huggingface.co/datasets/ucberkeley-dlab/measuring-hate-speech} \\
Multitarget-CONAN & Research Only & \href{https://github.com/marcoguerini/CONAN}{https://github.com/marcoguerini/CONAN} \\
NewsHate & CC BY & \href{https://ojs.aaai.org/index.php/ICWSM/article/view/15028/14878}{https://ojs.aaai.org/index.php/ICWSM/article/ view/15028/14878}\\
ReligiousHate & MIT & \href{https://github.com/dhfbk/religious-hate-speech}{https://github.com/dhfbk/religious-hate-speech}\\
SlurCorpus & MIT & \href{https://github.com/networkdynamics/slur-corpus}{https://github.com/networkdynamics/slur-corpus}\\
Stormfront & CC BY-SA 3.0 ES & \href{https://github.com/Vicomtech/hate-speech-dataset}{https://github.com/Vicomtech/hate-speech-dataset}\\
SWAD & CC BY NC & \href{https://aclanthology.org/2020.lrec-1.765.pdf}{https://aclanthology.org/2020.lrec-1.765.pdf}  \\
ToxiCR & GPL-3.0 license & \href{https://github.com/WSU-SEAL/ToxiCR}{https://github.com/WSU-SEAL/ToxiCR}\\
TwitterCRT & CC BY 4.0 & \href{http://github.com/zeerakw/hatespeech}{http://github.com/zeerakw/hatespeech}\\
TwitterExpert & CC BY 4.0 & \href{http://github.com/zeerakw/hatespeech}{http://github.com/zeerakw/hatespeech} \\
TwitterSA & Unknown & \href{https://www.kaggle.com/datasets/arkhoshghalb/twitter-sentiment-analysis-hatred-speech}{https://www.kaggle.com/datasets/arkhoshghalb/twitter-sentiment-analysis-hatred-speech}\\
USElect & CC BY 4.0 & \href{https://www.ims.uni-stuttgart.de/forschung/ressourcen/korpora/stance-hof/}{https://www.ims.uni-stuttgart.de/forschung/ressourcen/korpora/stance-hof/}\\ 
AMI18 & CC BY-NC 4.0 & \href{https://github.com/MIND-Lab/Automatic-Misogyny-Identification}{https://github.com/MIND-Lab/Automatic-Misogyny-Identification}\\ 
EDOS & CC0-1.0 & \href{https://github.com/rewire-online/EDOS}{https://github.com/rewire-online/EDOS}\\
HASOC19 & Open Access & \href{https://hasocfire.github.io/hasoc/2019/call_for_participation.html}{https://hasocfire.github.io/hasoc/2019/call\_for\_participation.html}\\
HatEval19 & CC BY 4.0 & \href{https://github.com/msang/hateval/tree/master/SemEval2019-Task5}{https://github.com/msang/hateval/tree/master/ SemEval2019-Task5} \\
HateXplain & CC BY 4.0 & \href{https://huggingface.co/datasets/Hate-speech-CNERG/hatexplain}{https://huggingface.co/datasets/Hate-speech-CNERG/hatexplain} \\
Jigsaw & CC0-1.0 & \href{https://www.kaggle.com/competitions/jigsaw-unintended-bias-in-toxicity-classification/data}{https://www.kaggle.com/competitions/jigsaw-unintended-bias-in-toxicity-classification/data} \\
OffensEval20 & CC BY 4.0 & \href{https://sites.google.com/site/offensevalsharedtask/offenseval-2019}{https://sites.google.com/site/offensevalsharedtask/offenseval-2019}\\
ParaDetox & OpenRail++ & \href{https://huggingface.co/datasets/s-nlp/paradetox}{https://huggingface.co/datasets/s-nlp/paradetox}\\
ToxicSpans & CC0-1.0 & \href{https://huggingface.co/datasets/heegyu/toxic-spans}{https://huggingface.co/datasets/heegyu/toxic-spans}\\ 
\bottomrule
\end{tabular}
\caption{Overview of the licenses associated with each resource utilized in this work for experiments.}
\label{tab:overview-license}
\end{table*}

The licenses associated with the datasets utilized in this study are consistent with the intended use of conducting academic research on various NLP application for positive impact.

As all the licenses support public access to the data, we will opensource our final combined instruction training dataset with the most strict license that will support only research purposes that serve responsible usage of datasets and the model for social good.

\section{Datasets Examples}
\label{sec:app_datasets_examples}

Here, we provide several examples of the texts and the corresponding labels from the used datasets for the classification task (see Table~\ref{tab:app_examples}) as well as datasets used for the hate speech mitigation with generation tasks (see Table~\ref{tab:app_examples_generation}).

\textcolor{red}{\textit{Warning: The examples are included solely to illustrate the nature of the dataset and are not intended to offend or target any individual or group.}}

\begin{table*}[ht!]
    \centering
    \begin{tabular}{p{2cm}p{9cm}c}
    \toprule
        \multicolumn{1}{c}{\textbf{Dataset}} & \multicolumn{1}{c}{\textbf{Examples}} & \textbf{Label}\\
        \midrule
         \multirow{7}{*}[-1.5ex]{\makecell{\textsc{AMI18} \\ \textsc{Subtask 1}}}& There are some wins that will make u go hysterical. Some makes you happy but same time you are sad & \multirow{2}{*}{Non-misogyny} \\ \cmidrule{2-3}
          & Shut the f*ck up you f*cking monkey jew b*tch c*nt n*gger d*ck l*cking & \multirow{2}{*}{Misogyny} \\ \cmidrule{2-3}
         & he thing is if a woman speaks out about rape on Nigerian twitter.The backlash would be shocking. It's not good for the mind. Hence silence & \multirow{3}{*}{Non-misogyny} \\
        \midrule

         \multirow{8}{*}[-1.5ex]{\makecell{\textsc{EDOS} \\ \textsc{Subtask 2}}}& MRA stuff like men saying they wished they hadn't had daughters, calling ALL American women wh*res, etc. Evidence if this? & \multirow{3}{*}{Threats} \\ \cmidrule{2-3}
         & Is this ASD (anti sl*t defence) or what? How do I prevent this? Spend more time with her before and after. & \multirow{2}{*}{Derogation} \\ \cmidrule{2-3}
         & Not our fault she chose to date an abusive sc*mbag. & Animosity \\ \cmidrule{2-3}
         & Someone k*ll her, really, I'm not kidding, someone c*t her throat & \multirow{2}{*}{Prejudiced Discussion} \\
         \midrule

         \multirow{5}{*}[-1.5ex]{\makecell{\textsc{HatEval19} \\ \textsc{Subtask 1}}}& And people still want to let them all in?!? \#BuildTheDamnWall \#NoDACA \#ICE \#LockThemAllUp \#SendThemBack & \multirow{2}{*}{Hateful} \\ \cmidrule{2-3}
         & It's time for Africans to reevaluate their relations with \#China.  & \multirow{2}{*}{Non-hateful} \\ \cmidrule{2-3}
         & Bitch shut up u a whore & Hateful \\

        \bottomrule
    \end{tabular}
    \caption{Examples from a few classification datasets used in our work for experiments. We included multiple subtasks and domains, covering just high-level classification of hate/non-hate as well as other more fine-grained labels.}
    \label{tab:app_examples}
\end{table*}

\begin{table*}[h!]
    \centering
    \begin{tabular}{p{2cm}p{11cm}}
    \toprule
        \multicolumn{1}{c}{\textbf{Dataset}} & \multicolumn{1}{c}{\textbf{Examples}} \\
        \midrule
         \multirow{7}{*}[-1ex]{\textsc{ToxicSpans}}& \textbf{Input Text}: your an id*ot, this is a tax based on a lie \newline \textbf{Toxic Span}: id*ot \\ \cmidrule{2-2}
         & \textbf{Input Text}: Have you walked around Denver lately , or Civic Center Park ? These poor homeless people are p*gs ! \newline \textbf{Toxic Span}: p*gs \\ \cmidrule{2-2}
        & \textbf{Input Text}: M*ron.  Illegals have no Constitutional rights \newline \textbf{Toxic Span}: M*ron \\

        \midrule

         \multirow{6}{*}[-1ex]{\textsc{ParaDetox}}& \textbf{Input Text}: the police dont have time for this cr*p. \newline \textbf{Detoxified}: The police have no time for this. \\ \cmidrule{2-2}
         & \textbf{Input Text}: its a crock of sh*t , and you know it . \newline \textbf{Detoxified}: It's senseless, you know \\ \cmidrule{2-2}
        & \textbf{Input Text}: i didnt vote for the liar \newline \textbf{Detoxified}: I didn't vote for him \\
        
        \bottomrule
    \end{tabular}
    \caption{Examples from a few generation datasets used in our work for experiments. We included not only classification, but generation tasks showing that instruction-tuned model can be used for more variety of proactive ways for hate speech mitigation.}
    \label{tab:app_examples_generation}
\end{table*}

\section{Model Comparison} \label{app:models}
In this section, we compare varying models to understand their effect.
Our main model, based on \texttt{Qwen3-4B-Instruct-2507}, is a newer version of Qwen3, focused only on instruction tuning, rather than a mixture of instruction and reasoning data. We compare to the older Qwen3 model, simply named \texttt{Qwen3-4B}. We disable its reasoning capabilities. To better distinguish between the two, we refer to our main model as HIPPO and the older model version as Qwen3. We also compare to \texttt{Llama3.2-3B-Instruct}, and \texttt{Phi4-Mini-Instruct} in Table \ref{tab:model_types}.

\begin{table*}[!htp]\centering
\begin{tabular}{lrrrrr}\toprule
Dataset &Llama3.2 &Phi4-mini &Qwen3 &HIPPO \\\midrule
AMI18 - A &78.7 &75.7 &79.2 &\textbf{81.0} \\
AMI18 - B &70.3 &54.8 &72.5 &\textbf{73.4} \\
EDOS - A &83.5 &83.5 &\textbf{84.6} &83.7 \\
EDOS - B &70.0 &61.9 &67.5 &\textbf{71.2} \\
EDOS - C &43.8 &41.4 &47.5 &\textbf{51.8} \\
HASOC19 - A &77.6 &74.5 &\textbf{78.1} &76.7 \\
HASOC19 - B &\textbf{62.2} &60.4 &56.6 &58.5 \\
HASOC19 - C &52.7 &\textbf{53.1} &50.4 &52.8 \\
HatEval19 - A &55.2 &56.9 &\textbf{57.6} &56.8 \\
HatEval19 - B &77.4 &79.8 &79.8 &\textbf{80.0} \\
HateXplain &68.8 &66.2 &\textbf{70.2} &69.5 \\
Jigsaw &80.6 &79.9 &80.1 &\textbf{80.7} \\
OffensEval20 - A &92.6 &\textbf{93.1} &92.6 &92.8 \\
OffensEval20 - B &65.9 &\textbf{73.0} &69.0 &66.2 \\
OffensEval20 - C &70.9 &54.7 &\textbf{72.0} &70.3 \\
ParaDetox &\textbf{69.4} &68.0 &64.9 &64.8 \\ 
ToxicSpans &62.6 &\textbf{65.1} &63.3 &64.3 \\
\midrule
Average &69.5 &67.2 &69.8 &\textbf{70.3} \\
\bottomrule
\end{tabular}
\caption{Comparison of model types. We can observe that Qwen3 family models (\texttt{Qwen3} and \texttt{HIPPO}) perform better and more stable in comparison to other same-sized models from other families.}\label{tab:model_types}

\end{table*}

Here, we see that while HIPPO performs best, and both Qwen-based models perform better overall, the other models show similar performance, demonstrating the effectiveness of our training setup on multiple architectures.

\section{Extended results}
\label{app:extended_results}

\begin{figure*}[h!]
    \centering
    \includegraphics[width=\linewidth]{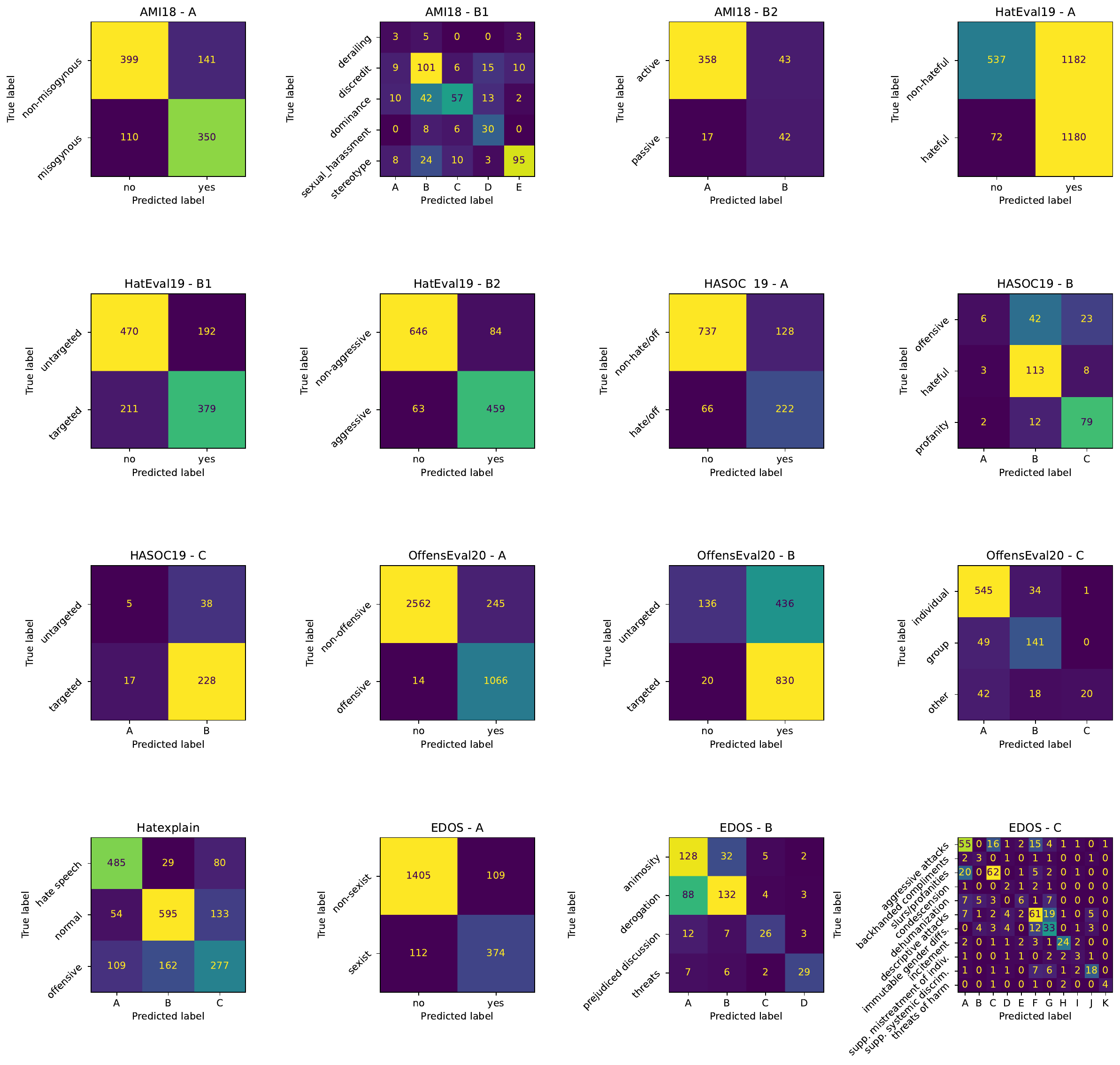}
    \caption{Confusion matrices for all classification-based tasks in the test set. Predicted label categories correspond to the true label categories, abbreviated for brevity.}
    \label{fig:conf_mtx}
\end{figure*}

In Figure \ref{fig:conf_mtx}, we show the confusion matrices for every classification-based task tested (for English) for our unified model. 

Overall, our unified model demonstrates strong performance across a range of labeling schemes, quite accurately predicting binary categories from diverse domains---such as hate vs./non-hate, misogynous/non-misogynous, and offensive/non-offensive---with minimal label confusion. It also performs effectively on more fine-grained classification tasks, including \texttt{AMI18-Task B2} and \texttt{EDOS-Task C}, which distinguish between categories such as slurs, attacks, threats, harassment, and others. At the same time, the model occasionally confuses targeted and untargeted content and, in some tasks, hateful language with just offensive ones. Nevertheless, these detailed per-label results indicate that a single LLM tuned on the heterogeneous harmful content detection tasks can indeed predict beyond binary labels and generalize across multiple domains, task formulations, and categories granularities.

\begin{figure*}[h!]
    \centering
    \includegraphics[width=\linewidth]{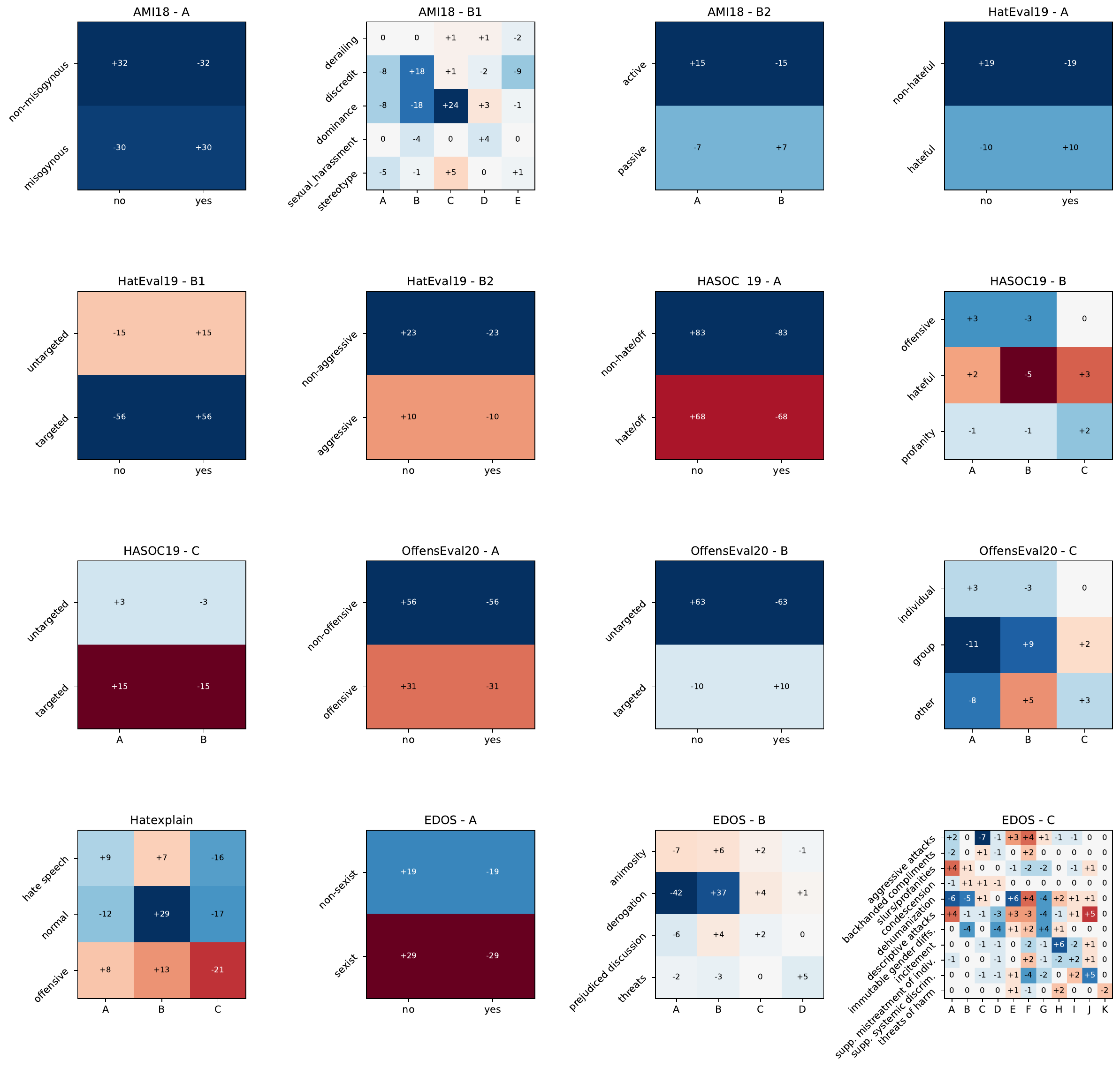}
    \caption{Difference of confusion matrices between the unified model and individual models. Blue indicates better performance for unified, red vice versa.}
    \label{fig:conf_mtx_diff}
\end{figure*}

We compare the unified model against the individual models in Figure \ref{fig:conf_mtx_diff}. From these, we can see that, the results on AMI18-B and EDOS-B improve with the unified model due to less confusion of the categories \textit{discredit} and \textit{dominance} in AMI18, and \textit{derogation} and \textit{animosity} in EDOS. There is a high overlap in \textit{discredit} and \textit{derogation} in their definition (as defined by the shared tasks), and similarly for \textit{animosity} and \textit{dominance}, so the unified model is able to benefit from the similarities of these two tasks, while the individual models are unable to do so.

\section{Further Discussion} \label{app:discussion}
\paragraph{Label Disagreement}
Unifying hate speech datasets has an underlying issue: label disagreement. While we did handle this in our filtering step for exact duplicates, sentences with similar sentiment are more difficult to identify, especially in the context of hate speech, where subtle differences can lead to different classifications. \citet{fortuna2020toxic} noted that label disagreement is a major inhibitor in unifying datasets. Annotator guidelines are a contributing factor, with different definitions of hate speech and inconsistent categorizations that do not overlap across datasets. Annotators themselves can also have disagreements that lead to inconsistencies within a dataset.

As we noted in our results in Section \ref{subsect:xtask}, differing distributions among datasets could be a result of different guidelines, and can hurt the cross-task performance if the model has not been trained on that specific dataset. We note that in the HatEval19-C task of determining whether a tweet is targeted, the guidelines specify it should target an individual only, not a group, while other datasets include groups (e.g., HASOC19). 

Including definitions and guidelines within a prompt could be a solution to this, however it also makes it difficult to prevent overfitting. If the guidelines and definitions are always the same, it will become a signal for that specific dataset. A future approach may be to paraphrase these guidelines for increased variety. Nevertheless, our joint model outperforms individually trained models, so the benefits of joint training overall outweigh the drawbacks.


\paragraph{Data Contamination} 
Data contamination is a widespread issue for closed-source LLMs, and it is difficult to determine the level of contamination, given that we have no access to the datasets used in training. It is important to distinguish the differences in types of contamination, particularly \textit{text contamination} versus \textit{text-label contamination} \cite{jiang2024investigating}. As the majority of the datasets we use originate from public social media such as Twitter, it is highly likely that there is text contamination. However, this is also the case for a number of the best-reported models, as they are additionally trained on unlabeled data. For example, works on OffensEval20 and EDOS included a large amount of unlabeled data for training, as the data was provided by the shared task organizers. 

Concerning text-label contamination, it is more difficult to assess whether models have been contaminated in such a way. We are confident that at least our results on ParaDetox are not contaminated, as the test sets were obtained from the organizers and are not freely available online.

To the extent that contamination affects our results, we can be confident that it does not affect RQs 2, 3, and 4, as these experiments all use the same base model. For RQ1, we can be sure that our unified method outperforms individual training, and our comparison to GPT5-mini should also be fair, as there is no reason to expect GPT5-mini to be less contaminated than Qwen3. Overall, while data contamination may inflate performance on some tasks, instruction tuning on hate speech data still remains a promising method for hate speech processing, clearly outperforming simple prompting, and demonstrating a level of cross-task generalization that is otherwise impossible with BERT-style models.

\section{Usage of AI Assistants}
In this paper, an AI assistant was utilized in the writing process. ChatGPT was used only for paraphrasing and checking the grammar in some paragraphs. 
During the development phase, Cursor was used for code completion and assistance.
    
\end{document}